\documentclass{article}

\usepackage{url}            
\usepackage{booktabs}       
\usepackage{amsfonts}       
\usepackage{nicefrac}       
\usepackage{microtype}      

\usepackage{algorithm}
\usepackage{algorithmic}
\usepackage{graphicx}
\usepackage{subcaption}
\usepackage{amsmath}
\usepackage{amsthm}
\usepackage{amssymb}
\usepackage{mathtools}
\usepackage{natbib}
\setcitestyle{numbers,square}
\usepackage{thmtools}
\usepackage{thm-restate}

\usepackage{diagbox}
\usepackage{multirow}
\usepackage{xcolor}
\usepackage[preprint]{neurips_2024}
\usepackage{hyperref}

\newcommand{\bx}{\boldsymbol{x}}

\newcommand{\bepsilon}{\boldsymbol{\epsilon}}
\newcommand{\beps}{\boldsymbol{\epsilon}}

\newcommand{\btheta}{\boldsymbol{\theta}}

\newcommand{\bI}{\boldsymbol{I}}

\newcommand{\bbR}{\mathbb{R}}

\newtheorem{remark}{\textbf{Remark}}

\newcommand{\mE}{\mathbb{E}}

\newcommand{\cS}{\mathcal{S}}
\newcommand{\cC}{\mathcal{C}}
\newcommand{\cN}{\mathcal{N}}

\newcommand{\cT}{\mathcal{T}}

\newcommand{\baralpha}{\bar{\alpha}}

\begin{document}
	
	\title{Enhancing Text-to-Image Editing via Hybrid Mask-Informed Fusion} 
		\author{Aoxue Li$^{1}$\thanks{equal contribution}, Mingyang Yi$^{1*}\thanks{corresponding author}$, Zhenguo Li\\
		$^{1}$ Huawei Noah's Ark Lab\\
		\texttt{\{yimingyang2,liaoxue2,li.zhenguo\}@huawei.com}
	}
	\maketitle
	
 	\begin{figure}[h]
		\centering
		\vspace{-0.4in}
		\includegraphics[scale=0.43]{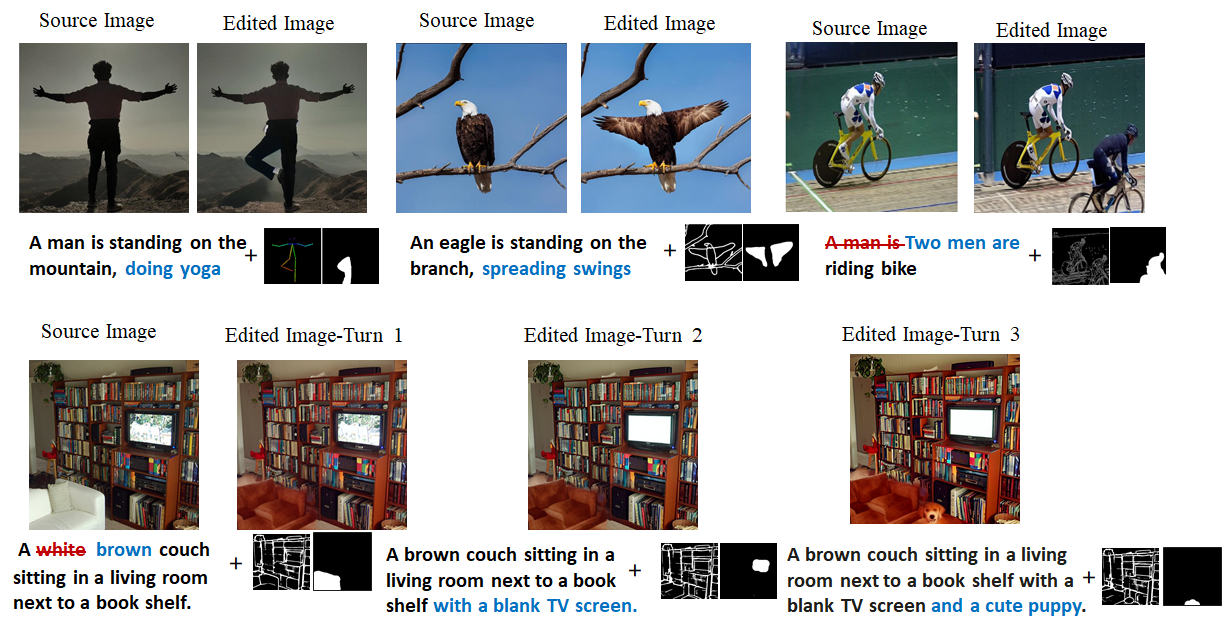}
		\caption{Our training-free methods on text-to-image editing on real images. The method integrates some human annotation e.g., (sketches as in text-to-image adapter \cite{mou2023t2i} and restricted editing region).}
		 \vspace{-0.1in}\label{fig:start_examples}
	\end{figure}
 
	\begin{abstract}
		Recently, text-to-image (T2I) editing has been greatly pushed forward by applying diffusion models. Despite the visual promise of the generated images, inconsistencies with the expected textual prompt remain prevalent. This paper aims to systematically improve the text-guided image editing techniques based on diffusion models, by addressing their limitations. Notably, the common idea in diffusion-based editing firstly reconstructs the source image via inversion techniques e.g., DDIM Inversion. Then following a fusion process that carefully integrates the source intermediate (hidden) states (obtained by inversion) with the ones of the target image. Unfortunately, such a standard pipeline fails in many cases due to the interference of texture retention and the new characters creation in some regions. To mitigate this, we incorporate human annotation as an external knowledge to confine editing within a ``Mask-informed'' region. Then we carefully Fuse the edited image with the source image and a constructed intermediate image within the model's Self-Attention module. Extensive empirical results demonstrate the proposed ``MaSaFusion'' significantly improves the existing T2I editing techniques.
	\end{abstract}
 
	\section{Introduction}\label{sec:Introduction}
	The field of Text-to-Image (T2I) generation has witnessed significant advancements through the development of large-scale diffusion probabilistic generative models, e.g., Stable Diffusion \cite{rombach2022high}. These powerful models create high-quality images that align with the provided text prompts \citep{radford2021learning}. Additionally, these T2I models are further applied as backbones for image editing tasks. 	During editing, compared with the source image, the ideal target image alters specific features to be consistent with the target text prompt, while maintaining the other features invariant \cite{mokady2023null}. In contrast to the training-based editing methods \cite{zhang2023adding,zhang2023hive,brooks2023instructpix2pix}, the training-free T2I editing methods are more tractable  \cite{meng2021sdedit,hertz2022prompt,mokady2023null,cao2023masactrl,tumanyan2023plug} due to their convenience in practice. However, these methods often fall short of practical needs, with many instances of failures that lead to unsatisfactory visual results \cite{mokady2023null}.
	\par
\begin{figure}[t!]
		\centering
		\vspace{-0.6in}
		\includegraphics[scale=0.4]{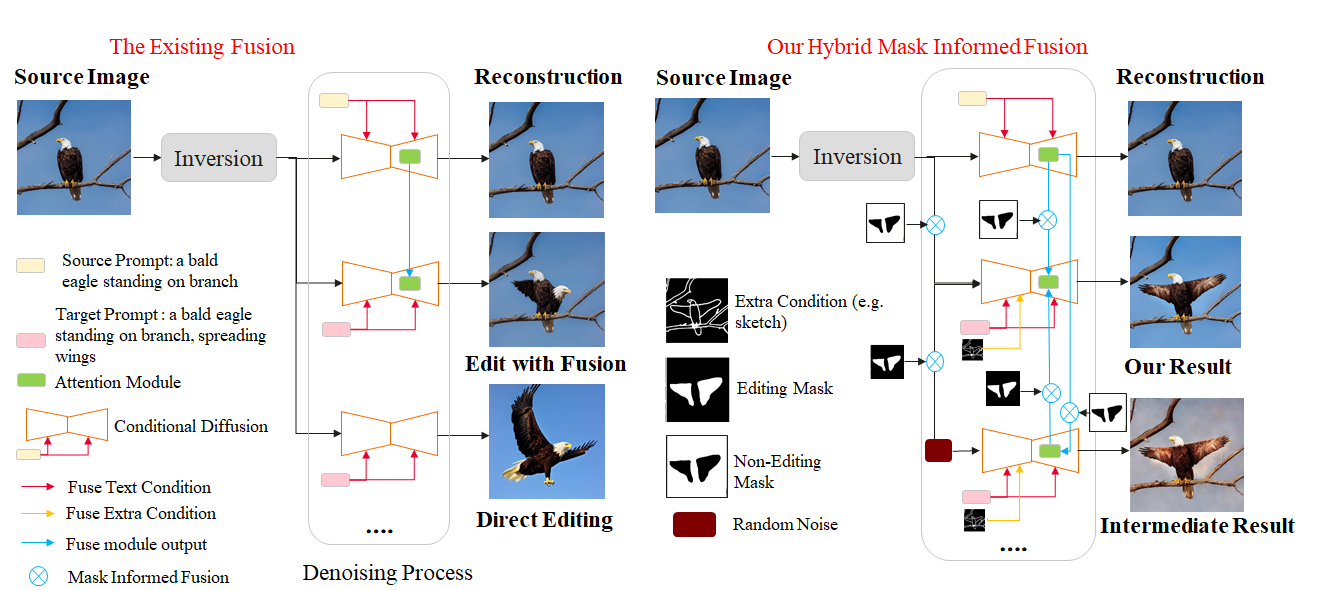}
		\caption{The comparison of the existing (left) and our (right) fusion processes (e.g. \cite{hertz2022prompt}) of generating desired target image. The standard pipeline is inversing the source image, and then fusing it with the target one. For the existing method on the left, the target image can be quite different from the source one. Thus, the fusion process between them is different in practice. We propose to first generate an \emph{intermediate image} with the desired shape under external conditions (e.g. T2I Adapter \cite{mou2023t2i}), then obtain the target image by fusing it with the source and intermediate images, depending on the pixels' relative position to a prior given editing region.}
		\vspace{-0.2in}
		\label{fig:fusion process}
	\end{figure}

	The goal of this paper is to enhance the current editing techniques to make them meet practical requirements. Our analysis begins with an examination of the failures in existing methods. We find that the failures are usually caused by the interference between preserving existing characteristics and generating new ones (as in Figure \ref{fig:fusion process}). To make this clear, we notice that the common idea in diffusion-based editing consists of two steps i.e., ``Inversion then Fusion'' \cite{hertz2022prompt,mokady2023null}. That says we first apply inversion techniques on the source image (e.g., DDIM Inversion \cite{song2020denoising}, Null-Text-Inversion \cite{mokady2023null}) to get its intermediate hidden states (from image to noise). Then, fusing these intermediate hidden states with the ones of the target image during its denoising (generation) process to make it similar to the source image. However, during the fusion process, the generation of variant features may occur (especially for the editing with variant shapes of objects) in mismatched positions relative to the source image, leading to confusion between preservation and generation. This confusion can result in a dominance of one mode in these positions, and through the neural network's attention mechanism \cite{vaswani2017attention}, potentially propagates the dominance throughout the entire image. 
	\par
	To counter this interference, we propose using external human annotations to define the boundaries of the edited area, thereby limiting both the editing itself and the possible spread of interference. Additionally, to make the fusion process easier, we create an intermediate image that has a consistent shape (while potentially different textures) with the target image. Concretely, we incorporate further external factors (e.g., the sketch of a desired target image) to produce such an intermediate image with a strong conditional generative model i.e., T2I Adapter \cite{mou2023t2i,cao2023masactrl}. Notably, the correlation between pixels (self-attention map \cite{vaswani2017attention}) of such intermediate image is potentially consistent with the ideal target image, as they have the same shape, while such correlation may vary from the source to target images. Inspired by this, we propose to fuse the final target image with the intermediate ones in the self-attention module of model \cite{vaswani2017attention,ronneberger2015u} for pixels in given edited regions, to generate desired varied characteristics. Oppositely, the source image is applied to fuse with the target ones for pixels out of the given edited region so that keep unedited pixels invariant.  
	By doing so, we separate the preservation and generation during editing, thus alleviating the interference between them. We term the proposed method as ``Masked region informed Self-Attention Fusion'' abbreviated as \textbf{\emph{MaSaFusion}}.      
	
	\par
	Empirical results of real-world image editing demonstrate the effectiveness of our proposed MaSaFusion. Concretely, we evaluate our approach on MagicBrush~\cite{zhang2023magicbrush}, which is a benchmark dataset in T2I editing with expertise in constructed ground-truth source-target image pairs. The quantitative results on the alignments of generated images with target images and text prompts show that the proposed MaSaFusion has superior performance, compared with the existing methods.
    
	\section{Related Work}\label{sec:related work}
	\paragraph{Text-to-Image Generation.} As the backbone technique of T2I editing, the goal of T2I generation is generating images that are aligned with provided textual prompts through a multi-modal vision-language model. The earlier works rely on the auto-regressive conditional prediction \cite{yu2022scaling,ramesh2021zero} or generative adversarial networks \cite{goodfellow2014generative,zhang2017stackgan,zhang2018stackgan++,zhang2021cross,kang2023scaling}, yet often resulted in images of suboptimal quality or misaligned with the textual prompts. Fortunately, the recent advancements in diffusion models have significantly advanced the field  \cite{ho2020denoising,song2020score,yi2023generalization,xue2023sa}. By combing the multi-modal vision-language model e.g., CLIP \cite{radford2021learning}, a succession of studies have attained state-of-the-art results in T2I generation. Notably, GLIDE \cite{nichol2022glide}, VQ-Diffusion \cite{gu2022vector}, DALL.E 2 \cite{ramesh2022hierarchical}, Imagen \cite{saharia2022photorealistic}, and Stable-Diffusion \cite{rombach2022high} are the most successful among them up till now. In our research, we leverage the Stable Diffusion as the backbone model, due to its superior performance and practicality in editing applications.
 
	\paragraph{Text-to-Image Editing.} As we have claimed, with the advancements in foundational techniques (T2I generation), T2I editing has made substantial progress. The methodologies can be categorized into training-based or training-free methods, depending on whether an extra fine-tuning stage is required before the editing process. Considering the extra fine-tuning stage potentially hind the application of training-based methods (e.g., InstructP2P \cite{brooks2023instructpix2pix}, Paint by Example \cite{yang2023paint}, SINE \cite{zhang2023sine}, ControlNet \cite{zhang2023adding} or HIVE \cite{zhang2023hive}), our research prioritizes training-free approaches, which offer greater flexibility in terms of inference speed and computational resource demands. To date, the existing training-free techniques mainly include SDE-Diffedit \cite{couairon2022diffedit,nie2023theblessing}, Prompt-to-Prompt (P2P) \cite{hertz2022prompt}, Plug-and-Play (PnP) \cite{tumanyan2023plug}, and Masactrl \cite{cao2023masactrl}, where the first three methods are suitable to editing with invariant objects' shapes, and the last method is devised for objects' shapes variant editing. As clarified in Section \ref{sec:Introduction}, the common idea among these methods follows the pipeline  ``Inversion then Fusion''. However, as in Figure \ref{fig:fusion process}, these methods fail in many cases owing to inappropriate fusion processes. 
	\par

    \paragraph{Editing with External Knowledge.} As in this paper, much-existing literature introduce the external information e.g., the sketch map of the target image in T2I Adapter \cite{mou2023t2i} and ControlNet \cite{zhang2023adding}, the mask-informed editing region in Blended Diffusion \cite{Avrahami2023BlendedDiffusion} and GLIDE \cite{nichol2022glide}. According to their observations, the external information significantly improves the quality of edited images. Thus, during our MaSaFusion, we properly combine the induced external information in the T2I Adapter and mask-informed editing region and improve the existing external knowledge-induced methods.  
    
	\section{Preliminaries}\label{sec:Preliminaries}
	\paragraph{Latent Diffusion Model.} 
	This study examines the standard methodology of text-to-image synthesis with diffusion models \cite{ho2020denoising,sohl2015deep}, which crafts the desired image from Gaussian noise through a gradual denoising process. To do so, the model is learned to predict the noise that has been added to the data. Our focus is on the Stable (latent) Diffusion model \cite{rombach2022high},  wherein the diffusion process occurs within the latent space of an image, encoded by a Vector Quantized (VQ) model \cite{esser2021taming}). In practice, given the generated latent, the synthesized real-word image is then reconstructed by an image decoder as in \cite{esser2021taming}.  
	\par
	The process of diffusion model is: given data $\bx_{0}$ and condition $\cC$ (textual prompt embedding from CLIP encoder \cite{radford2021learning} in this paper) from target distribution, we construct noisy data $\bx_{t}$ as 
	\begin{equation}
		\small
		\bx_{t} = \sqrt{\baralpha_{t}}\bx_{0} + \sqrt{1 - \baralpha_{t}}\beps_{t},
	\end{equation} 
	with $0\leq t \leq T$, where $\beps_{t}\sim \cN(0, \bI)$ and is independent with $\bx_{0}$, $\baralpha_{t}$ monotonically decreases with $t$, and $\baralpha_{t}\to 0$ (resp. $\baralpha_{t}\to 1$) for $t\to 1$ (resp. $t\to T$). By minimizing 
	\begin{equation}
		\small
		\min_{\btheta}\mE\left[\left\|\beps_{\btheta}(t, \bx_{t}, \cC, \emptyset) - \beps_{t}\right\|^{2}\right],
	\end{equation}
	the model learns to predict the injected noise in $\bx_{t}$. It is worthy to note that the noise prediction $\beps_{\btheta}(t, \bx_{t}, \cC, \emptyset)$ is constructed by classifier-free guidance model \cite{ho2021classifier}, such that
	\begin{equation}\label{eq:noise prediction}
		\small
		\beps_{\btheta}(t, \bx_{t}, \cC, \emptyset) = w\tilde{\beps}_{\btheta}(t, \bx_{t}, \cC) + (1 - w)\tilde{\beps}_{\btheta}(t, \bx_{t}, \emptyset),
	\end{equation}
	where $\emptyset$ is CLIP embedding of null text `` '', $w$ is guidance scale, and $\tilde{\beps}_{\btheta}$ is a model e.g., UNet \cite{ronneberger2015u}. 
	\par
	Since $\bx_{T}\approx \cN(0, \bI)$, we substitute it with standard Gaussian, then implement the following deterministic DDIM sampling process to gradually remove the noise starting from a Gaussian $\bx_{T}$.
	\begin{equation}\label{eq:DDIM}
		\small
		\bx_{t - 1} = \sqrt{\frac{\baralpha_{t - 1}}{\baralpha_{t}}}\bx_{t} + \left(\sqrt{\frac{1 - \baralpha_{t - 1}}{\baralpha_{t - 1}}} - \sqrt{\frac{1 - \baralpha_{t}}{\baralpha_{t}}}\right)\beps_{\btheta}(t, \bx_{t}, \cC, \emptyset).
	\end{equation}
	Technically, the forward process of DDIM is a discretion of an ordinary differential equation (ODE) \cite{song2020denoising}. Then, owing to the reversibility of it, the following DDIM inversion \cite{song2020denoising,mokady2023null} process on $\bx_{0}$ gives the corresponding initial noise $\bx_{T}$ such that $\bx_{0}$ is reconstructed by DDIM \eqref{eq:DDIM}. 
	\begin{equation}\label{eq:DDIM inversion}
		\small
		\bx_{t} = \sqrt{\frac{\baralpha_{t}}{\baralpha_{t - 1}}}\bx_{t - 1} + \left(\sqrt{\frac{1 - \baralpha_{t}}{\baralpha_{t}}} - \sqrt{\frac{1 - \baralpha_{t - 1}}{\baralpha_{t - 1}}}\right)\beps_{\btheta}(t, \bx_{t - 1}, \cC, \emptyset).
	\end{equation}
	Interestingly, it has been found in \cite{mokady2023null} that running the DDIM inversion \eqref{eq:DDIM inversion} with source textual prompt involved in it can not reconstruct the ideal $\bx_{0}$. To tackle this, \cite{mokady2023null} proposes the technique called ``null-text inversion'', more details of this can be checked in Appendix \ref{app:null-text}. 
	\par
	In this paper, the intermediate states of source and target images are respectively represented as $\bx_{t}^{\cS}$ and $\bx_{t}^{\cT}$. The source and target prompt conditions are $\cC^{\cS}$ and $\cC^{\cT}$, respectively. 
	
	\paragraph{Blocks in Conditional UNet.} The standard network structure of conditional diffusion model is UNet \cite{ronneberger2015u,song2020score,ho2020denoising}, which consists of an encoder and another decoder structure with self-attention  \cite{vaswani2017attention} and cross-attention blocks in them, capturing the ``pixel-pixel'' and ``pixel-condition'' correlations, respectively. 
    The attention layer in UNet takes value of $(Q, K, V)$ with output $\mathrm{Attention}(Q, K, V) = \mathrm{softmax}(QK^{\top}/\sqrt{d})V$, and $d$ is the dimension of pixel-wise embedding. Here  
	\begin{equation}\label{eq:self-attent}
		\small
		Q = W_{Q}\phi(\bx_{t}); K = W_{K}\phi(\bx_{t}); V =  W_{V}\phi(\bx_{t}), 
	\end{equation}    
	for self-attention $\mathrm{Attention}_{SA}$, $\phi(\bx_{t})\in \bbR^{N\times d}$ is the flattened intermediate hidden states of $\bx_{t}$, and 
	\begin{equation}\label{eq:cross-attent}
		\small
		Q = W_{Q}\phi(\bx_{t}); K = W_{K}\cC; V =  W_{V}\cC 
	\end{equation}
	for cross-attention $\mathrm{Attention}_{CA}$. 
	
	\paragraph{Inversion then Fusion.} This is a common idea in diffusion based editing originated from \cite{hertz2022prompt}. Generally, the ideal output of target image keeps unedited characteristics invariant as in source image. To conduct this, one should inject the information from source image into the target one. Thus, a natural initial state of denoising process to generate target image is the ones obtained via inverting source image. However, as in \cite{hertz2022prompt} or Appendix \ref{sec:SIE} of this paper, the information from intermediate states $\bx_{t}^{\cS}$ should be injected into the denoising process of $\bx_{t}^{\cT}$ to guarantee the similarity between source and target images. For example, in \cite{hertz2022prompt}, the cross-attention module $\mathrm{Attention}_{CA}(Q^{\cT}, K^{\cT}, V^{\cT})$ of generating $\bx_{t}^{\cT}$ for each $t$ is substituted by $\mathrm{Attention}_{CA}(Q^{\cS}, K^{\cS}, V^{\cT})$, during denoising process. Here $(Q^{\cS}, K^{\cS}, V^{\cS})$ and $(Q^{\cT}, K^{\cT}, V^{\cT})$ are respectively the hidden states of generating $\bx_{t}^{\cS}$ and $\bx_{t}^{\cT}$.    
	
	\section{Method}
	\begin{figure}[t!]
		\centering
				\vspace{-0.6in}
		\includegraphics[scale=0.3]{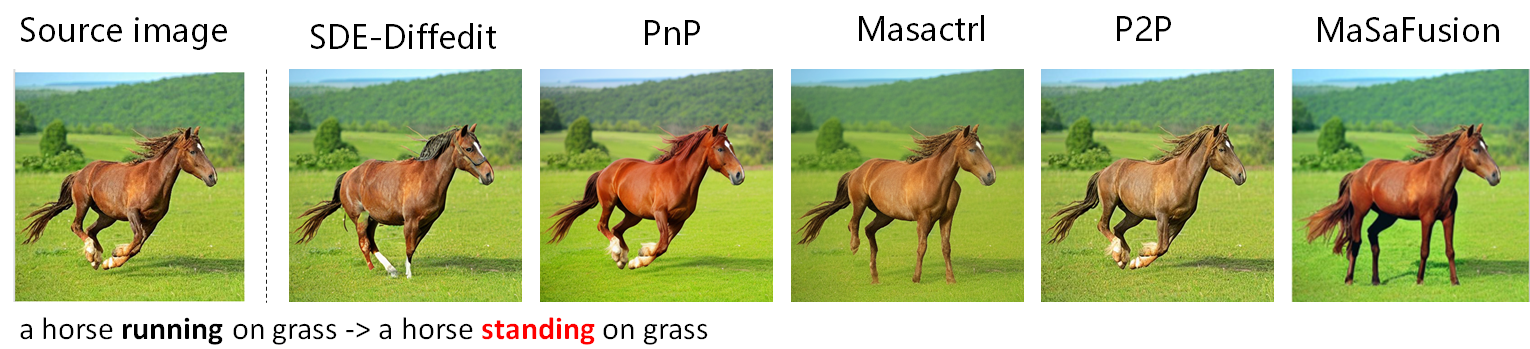}
				
		\caption{The typical failure cases of existing methods. The failures are mainly twofold, reconstructing source image (PnP, P2P) or generating .} \vspace{-0.2in}
		\label{fig:sve_example}
	\end{figure}
	In this section, we provide our method to tackle the image editing problem. We start our analysis with the failure cases in the existing methods. Following our analysis, we speculate that fusing an image close to the target one may improve the editing quality. To get such a close intermediate image, we incorporate extra human annotation required in T2I-Adapter \cite{mou2023t2i} (e.g., sketch of target image). Next, we leverage this intermediate image during the fusion process and restrict the editing in a prior known annotated region to further improve the generation of the target image.    
	\subsection{Interference During Fusion}\label{sec:interference When Doing SVE}
	We start with the failures of existing methods following ``inversion then fusion'' in Figure \ref{fig:sve_example}, which requires altering a ``running horse'' into a ``standing horse''. As can be seen, the failures are two-fold. 1): Reconstructing the source image, no desired features are generated e.g., PnP, P2P; 2): Generating image consistent with target prompt, but undesired features are generated e.g., SDE-Diffedit. 
	\par
	As claimed in Section \ref{sec:Preliminaries}, the idea of existing methods is based on making a fusion of the source image and a generated slightly related image, by manipulating the intermediate hidden states of the source and target images. However, this fusion process is particularly challenging, especially for the tasks with objects' shapes being variant (e.g., in Figure \ref{fig:sve_example}). Because, for such tasks, the requirement of generating images to have differing shapes, leading to significant interference in certain regions, during the fusion with the source image. This interference stems from the competing demands of the target prompt, which encourages the generation of new content, and the source image, which necessitates preservation. Striking a balance in these regions is complex and has led to failure cases as noted. To check this formally, we note that in both cross-attention and self-attention modules of UNet, the output ($N$-pixel values) is
	\begin{equation}
		\small
		\mathrm{Softmax}\left(\frac{QK^{\top}}{\sqrt{d}}\right)V = \left(\sum_{j}P_{1j}V_{j}, \cdots, \sum_{j}P_{Nj}V_{j}\right),
	\end{equation}
	where $\{P_{ij}\}$ in $\mathrm{Softmax}(QK^{\top} / \sqrt{d})$ is the attention map, and they are decided by the correlation between tokens in text prompt or pixels in image, respectively for cross-attention and self-attention modules. It is worth noting that the fusion process usually substitutes the attention map (cross-attention module in P2P \cite{hertz2022prompt}, self-attention module in MasaCtrl \cite{cao2023masactrl}) in the target image with the ones in the source image. However, if the objects' shapes are varied in the target images, both pixel-token level (cross-attention) and pixel-pixel (self-attention) correlations will vary from source to target images. This explains the failures of existing methods. 
	\begin{remark}
		For the editing with invariant objects' shapes, the existing methods e.g., P2P have relative better performances. However, when generating target prompt by textual encoder, its embedding of invariant tokens will vary, compared with the ones in source prompt. The variation of these unrelated tokens deteriorates the performance of existing methods. Thus we propose another simple method (does not require external information) tailored to such task, details are in Appendix \ref{sec:SIE}.  
	\end{remark}
	\par
	Based on our preceding analysis, we speculate that the attention map injected into the target image should be consistent with an underlying correct image for those regions to be edited. This speculation is further explored in Appendix \ref{app:SVE without Intermediate Images}. Nevertheless, such attention maps are unknown in practice. Consequently, we establish our first goal of editing: \textit{Obtaining an underlying correct attention map.} This aims to reduce interference during the fusion process.
	\subsection{Constructing Intermediate Image with Human Annotation}\label{sec:human annotation}
	Notably, the image that has a shape consistent with the underlying target image and correct textures (not necessarily consistent with the source image), is likely to have the same attention-map with the target image. Because the corresponding attention map creates a correct image with the desired shape. Thus, we propose to construct such an intermediate image to borrow its attention map in the fusion process of generating a target image.  
	To this end, we leverage the T2I-adapter \cite{mou2023t2i} which can generate the image with a shape consistent with the external signal \footnote{These signals are integrated into a pretrained stable diffusion model via injecting their encoded features into the hidden states of the model.} (e.g., a sketch of target image) \footnote{Please note that T2I-Adapter enables shape control instead of textures. So that it can not be directly applied to editing.}. Concretely, we leverage T2I-Adapter to generate an intermediate image with a shape consistent with the given external condition, under a target text prompt. To further guarantee the correct textures of such an intermediate image, we fuse it with the source image. 
	\par
	
	\begin{figure}[t!]
			\vspace{-0.5in}
		\centering
		\includegraphics[scale=0.4]{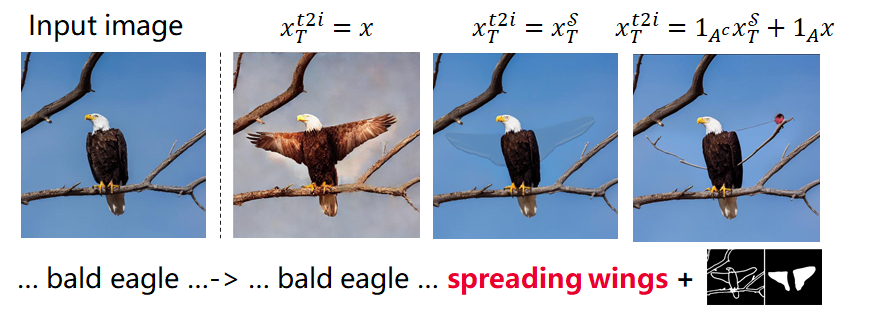}
		\caption{The generated \textbf{\emph{intermediate image}} with different initial noise $\bx_{T}^{\rm t2i}$ where $\bx_{T}^{\cS}$ is obtained by inversion, $\bx$ is a Gaussian noise, and $\textbf{1}_{A}$ is the indicator function on set $A$. The external condition (target image sketch) and shape of $A$ are also presented.}
		\label{fig:sve different initial noise}
	\end{figure}
	
	Additionally, the practical applications often involve prior knowledge of the regions that require editing, though these regions do not need to be defined with high precision. Consequently, we restrict the editing (as intermediate image fuses with the source ones) within these known, mask-informed regions, denoted as $A$. To do this, during the DDIM step \eqref{eq:DDIM} to generate the target image, we substitute the backbone model with pre-trained stable diffusion model in T2I-Adapter \cite{mou2023t2i}, and inject the external signal $\cC^{\mathrm{t2i}}$ as a conditional factor. The generated T2I-Adapter-based intermediate states are denoted as $\bx_{t}^{\rm t2i}$, and their attention maps are cached to be used in the fusion process of the target image. 
	\par
	To get $\bx_{t}^{\rm t2i}$ while fusing it with $\bx_{t}^{\cS}$, for its $i$-th pixel $\bx_{t}^{\rm t2i}(i)$, we revise the noise prediction as 
	\begin{equation}\label{eq:mask eps}
		\small
		\beps_{\btheta}(i) =
		\begin{cases}
			&\bar{\beps}_{\btheta}(t, \bx_{t}^{\rm t2i}, \cC^{\cS}, \cC^{\mathrm{t2i}}, \emptyset), \qquad i\notin A; \\
			&\beps_{\btheta}(t, \bx_{t}^{\rm t2i}, \cC^{\cT}, \cC^{\mathrm{t2i}}, \emptyset), \qquad i\in A.
		\end{cases}
	\end{equation}   
	We get $\bar{\beps}_{\btheta}(t, \bx_{t}^{\rm t2i}, \cC^{\cS}, \cC^{\rm{t2i}}, \emptyset)$ by substituting all the $\mathrm{Attention}_{SA}(Q^{\cT}, K^{\cT}, V^{\cT})$ as $\mathrm{Attention}_{SA}(Q^{\cS}, K^{\cS}, V^{\cS})$, for corresponded layers of model, with $\cC^{\mathrm{t2i}}$ injected. 
	By doing so, the pixels in $A^{c}$ are guaranteed to be similar with the source image, so that preserving correct features. On the other hand, the $\beps_{\btheta}(t, \bx_{t}^{\rm t2i}, \cC^{\cT}, \emptyset)$ in $A$ is used to generate desired variant features in the prior given editing region $A$. 

    	\begin{algorithm}[t!]
		\caption{MaSaFusion.}
		\label{alg:masafusion editing}
		\textbf{Input:} source prompt $\cS$, source image $\bx^{\cS}_{0}$, target prompt $\cT$, diffusion model $\beps_{\btheta}$, annotation mask region $A$, T2I-Adapter condition $\cC^{\mathrm{t2i}}$. 
		\begin{algorithmic}[1]
			\STATE  {\textbf{\emph{Inversion Step}}}
			\STATE  {Inversion (NTI Algorithm \ref{alg:null-text inversion} or DI Algorithm \ref{alg:direct inversion}) to generate $\bx^{\cS}_{T}$ and  intermediate hidden states $\{Q^{\cS}, K^{\cS}, V^{\cS}\}$ for each $t$;}
			\STATE  {\emph{\textbf{T2I-Adapter Intermediate Step}}}
			\STATE  {Initialize $\bx^{\mathrm{t2i}}_{T} = \bx$ with $\bx$ as a Gaussian noise;}
			\STATE  {DDIM \eqref{eq:DDIM} to generate $\bx^{\rm{t2i}}_{t}$ with $\bepsilon_{\btheta}$ in \eqref{eq:mask eps}, invoking annotation region $A$, T2I-Adapter condition $\cC^{\mathrm{t2i}}$, cached the hidden states $\{Q^{\mathrm{t2i}}, K^{\mathrm{t2i}}, V^{\mathrm{t2i}}\}$ for each $t$;}
			\STATE  {\emph{\textbf{Generating Target Image by Fusion}}, i.e., \emph{\textbf{fuse cached intermediate hidden states $\{Q^{\cS}, K^{\cS}, V^{\cS}\}$ and $\{Q^{\mathrm{t2i}}, K^{\mathrm{t2i}}, V^{\mathrm{t2i}}\}$}}}
			\STATE  {Initialize $\bx^{\cT}_{T} = \textbf{1}_{A^{c}}\bx_{T}^{\cS} + \textbf{1}_{A}\bx$;}
			\STATE  {
				DDIM \eqref{eq:DDIM} to generate $\bx_{t}^{\cT}$ with $\bepsilon_{\btheta}$ in \eqref{eq:eps masafusion}, by invoking annotation region $A$, T2I-Adapter condition $\cC^{\mathrm{t2i}}$;}
			\RETURN {Target image $\bx^{\cT}_{0}$.}
		\end{algorithmic}
	\end{algorithm}
    
    \begin{remark}
        Notably, the existing mask-based editing methods (e.g. GLIDE \cite{nichol2022glide} or Blended Diffusion \cite{Avrahami2023BlendedDiffusion}) reconstruct the unedited features out of editing regions $A^{c}$ by directly copying from the source image. Unlike theirs, we borrow the self-attentions from the intermediate hidden states of the source image to reconstruct unedited features. Compared with theirs, our method produces smoother and more natural features at the edge of editing regions. See Section \ref{sec:experiments} for more details.      
    \end{remark}
    
	\begin{remark}
		Our method necessitates specific external conditions and defined editing regions. The former (which is similarly used in MasaCtrl \cite{cao2023masactrl}) can be obtained as follows. We may first extract such conditions from the source image as in \cite{su2021pixel,mou2023t2i}. Then a simple handcrafted revision (e.g., revising several lines in sketch) on these conditions aligned with the target prompt yields the desired external conditions. As for the editing regions, they can be delineated by localizing the revised features within the external conditions. More details on these human annotations are in Section \ref{sec:ablation}.      
	\end{remark}
	\par
	In our experiments, we observed that the initial noise vector $\bx_{T}^{\rm t2i}$ significantly impacts the generated images. To see this, we generate the target image with revised noise prediction \eqref{eq:mask eps} under different initialized $\bx_{T}^{\rm t2i}$ in Figure \ref{fig:sve different initial noise}. As can be seen, expected for the data generated under $\bx_{T}^{\rm t2i} = \bx$, though the other two initialized $\bx_{T}^{\rm t2i}$ generate images consistent with the desired shape, their textures are incorrect. \footnote{In fact, when taking different $\bx$, initializing $\bx_{T}^{\cT}$ with $\textbf{1}_{A^{c}}\bx_{T}^{\cS} + \textbf{1}_{A}\bx$ occasionally generates desired intermediate image.} Therefore, we suggest to generate the intermediate image under $\bx_{T}^{\rm t2i} = \bx$. Besides that, we note that the two failure cases are originated from the incorrect self-attention map in $A$, as the pixels in $A$ are showed as inappropriate weighted sum of the other pixels. For example, the pixels under $\bx_{T}^{t2i} = \textbf{1}_{A^{c}}\bx_{T}^{\cS} + \textbf{1}_{A}\bx$ incorrectly pay attentions to branch and sky, instead of the eagle itself. This again illustrates the necessity of correct attention map. 
	\subsection{Mask Region Informed Self-Attention Fusion}
	
	With the correct attention-map from intermediate image, we propose to borrow it into the fusion process of generating target image as clarified before. To do this, we revise the noise prediction as of generating final target image as 
	\begin{equation}\label{eq:eps masafusion}
		\small
		\beps_{\btheta}(i) =
		\begin{cases}
			&\bar{\beps}_{\btheta}(t, \bx_{t}^{\cT}, \cC^{\cS}, \cC^{\mathrm{t2i}}, \emptyset), \qquad \qquad i\notin A; \\
			&\hat{\beps}_{\btheta}(t, \bx_{t}^{\cT}, \bx_{t}^{\mathrm{t2i}}, \cC^{\cT}, \cC^{\mathrm{t2i}}, \emptyset), \qquad i\in A,
		\end{cases}	
	\end{equation}
	where $\hat{\beps}_{\btheta}(t, \bx_{t}^{\cT}, \bx_{t}^{\mathrm{t2i}}, \cC^{\cT}, \cC^{\mathrm{t2i}}, \emptyset)$ is obtained via substituting the self attention module $\mathrm{Attention}_{SA}(Q^{\cT}, K^{\cT}, V^{\cT})$ as $\mathrm{Attention}_{SA}(Q^{\mathrm{t2i}}, K^{\mathrm{t2i}}, V^{\cT})$ (Algorithm \ref{alg:masafusion editing}) for corresponded layers of model. By doing so, the self-attention map of $\bx_{t}^{\cT}$ is exactly the ones of $\bx_{t}^{t2i}$. The $\bar{\beps}_{\btheta}$ is same as in \eqref{eq:mask eps}, which preserves the features from source image in unediting region $A^{c}$.  
	\par
	We summarize our method as Mask region informed Self-attention Fusion (MaSaFusion) in Algorithm \ref{alg:masafusion editing} and Figure \ref{fig:fusion process}. Please note that, we initialize $\bx_{T}^{\cT}$ with $\textbf{1}_{A^{c}}\bx_{T}^{\cS} + \textbf{1}_{A}\bx$, so that $\bx_{t}^{\cT}$ can generate new features in editing region $A$. The complete process of our method is summarized as firstly generating the intermediate image, then wisely fusing it and the source image with the desired target image.    
	
	\section{Experiments}\label{sec:experiments}
	\begin{figure}[t!]
			\vspace{-0.5in}
		\centering
		\includegraphics[scale=0.5]{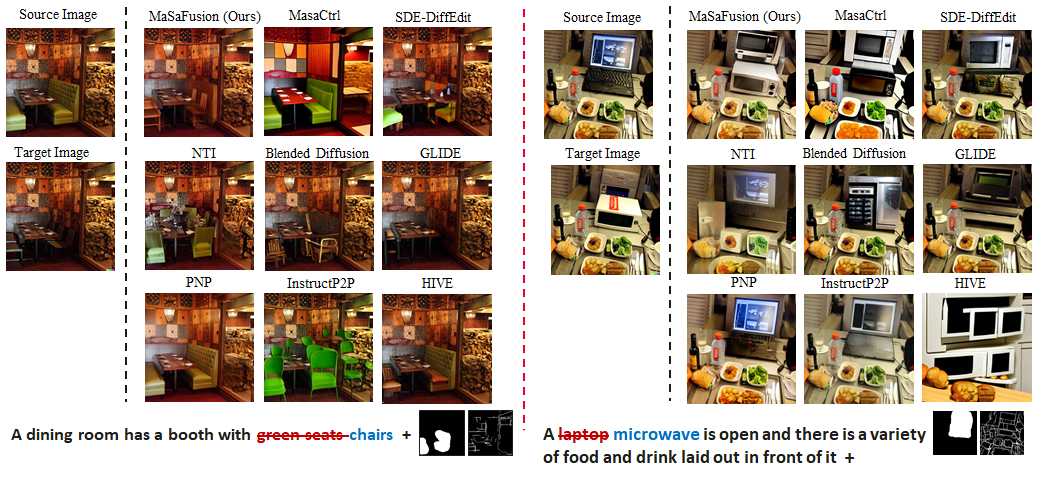}
		\caption{The edited images of existing methods and our MaSaFusion (NTI based). Here we present the results of single-turn editing, more results of multi-turn editing are in Appendix \ref{app:More Results on SVE Task}.}
         \vspace{-0.1in}
		\label{fig:sve_fusion}
	\end{figure}
	In this section, we empirically verify the effect of the proposed MaSaFusion in the task of T2I editing. 
	

    \subsection{Comparison with Existing Methods} 
    \paragraph{Dataset.} Magicbrush \cite{zhang2023magicbrush} is an expertise-constructed dataset consisting of many couples of source and human-evaluated edited target images. The editing includes many practical requirements e.g., object addition/replacement/removal, action changes, color/texture/pattern modifications. We evaluate our method and baseline methods in 1053 (including single and multi-turn editing) test data from Magicbrush by comparing the edited images with ground-truth target images provided by Magicbrush. 
    \par
    Notably, the used external conditions (sketch map of target images in T2I Adapter and masked region) are extracted from target images as in \cite{su2021pixel}. These external conditions are also used in the baseline methods with external information so that the comparisons with them are fair.

    \paragraph{Evaluation Metric.} We adopt the ones in \cite{zhang2023magicbrush}, where the alignments of obtained images to the ground-truth target images and the target text prompt are measured. Concretely, the $L_{1}$, $L_{2}$ distance between the generated and target images, DINO \citep{caron2021emerging}, Image-Image CLIP (CLIP-I) scores between the target and generated images, and Image-Text CLIP Scores (CLIP-T) between the generated images and target prompts \cite{radford2021learning}. 
    
	\paragraph{Setup.} We use Stable Diffusion \cite{ramesh2022hierarchical} model v1.4 integrated with T2I Adapter \cite{mou2023t2i} as backbone model. During inference stage (inversion \eqref{eq:DDIM inversion} or generation \eqref{eq:DDIM}), we use 50 steps DDIM \cite{song2020denoising}. For the inversion technique, our method can be combined with any proper ones. In this paper, we combine our MaSaFusion with null-text inversion (NTI) \cite{mokady2023null} and direct inversion (DI) \cite{ju2023direct}. It is worth noting that DI is computational efficient than NTI, i.e., 43s v.s. 184s for single image editing. However, the results in Table \ref{tbl:magicbrush} shows that MaSaFusion under NTI has slightly better result compared with combining it with DI. Therefore, a trade-off between efficiency and quality is existed when selecting inversion techniques (more details are in Appendix \ref{app:null-text}). 
\begin{figure}[t]
			\vspace{-0.5in}
		\centering
		\includegraphics[scale=0.35]{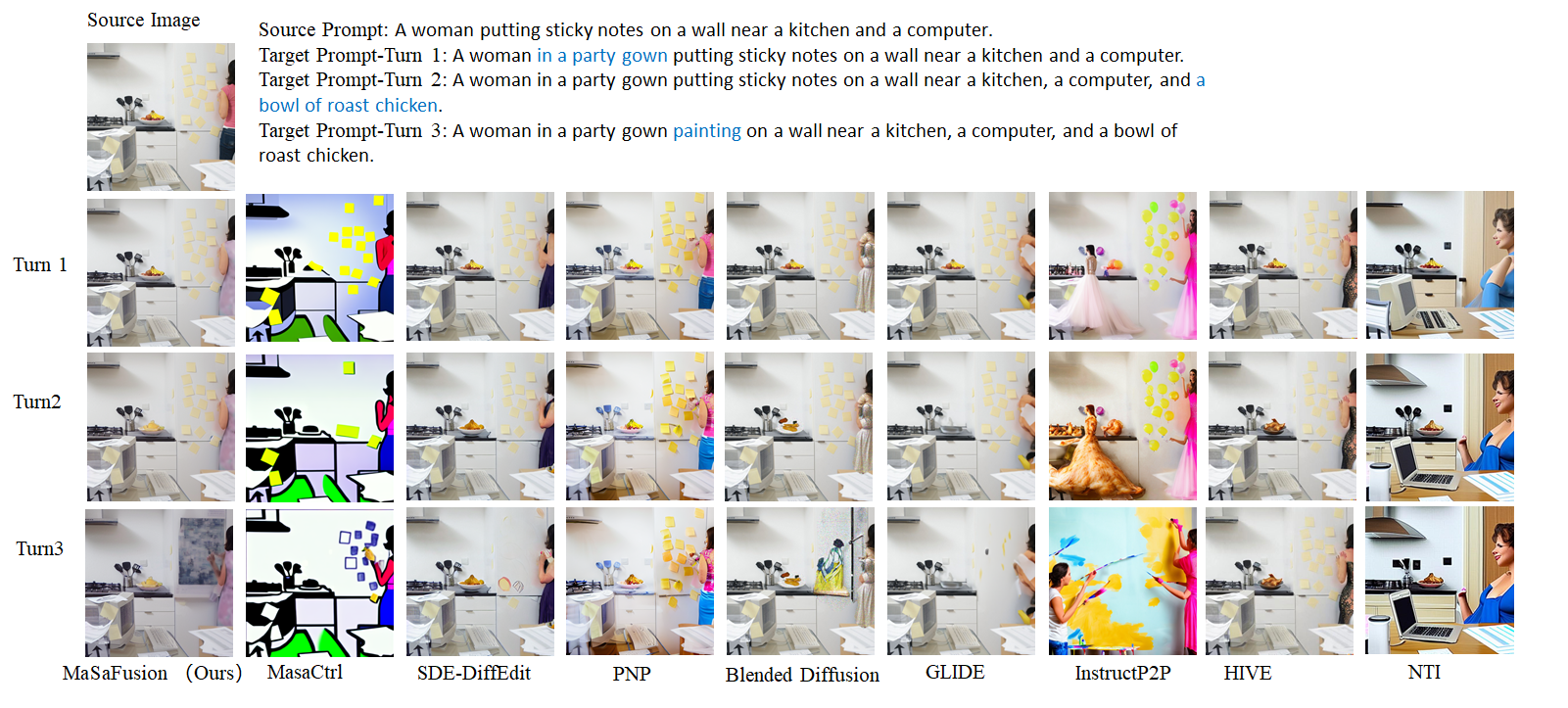}
		\caption{Multi-turn editing over different editing methods. The compared target images in Magicbrush are same for single-turn and multi-turn editing. The difference is that for multi-turn editing, the source image is set as the output of former editing.}
		\vspace{-0.1in}
		\label{fig:sve_fusion_multi_turn}
\end{figure}

\begin{table}[t]
\caption{The alignments with the target image/prompt of images generated under various T2I editing methods on MagicBrush dataset. We respectively compare the single-turn and multi-turn editing results of all methods with the same 1053 test data in MagicBrush. Please note that the multi-turn editing takes the previously edited images as source images, and the number of turns varies. The results with minimal average rank are marked in bold.}
\setlength{\tabcolsep}{1mm}
\label{tbl:magicbrush}
\centering
\scalebox{0.8}{
{
\begin{tabular}{llcccccc}\hline
{\bf Settings} &{\bf Methods}	&$L_1$($ \downarrow$) &$L_2$($\downarrow$) &CLIP-I($\uparrow$)&DINO($\uparrow$)    &CLIP-T($\uparrow$) & Ave Rank($\downarrow$) \\
\hline
\multirow{10}{*}{Single-Turn} 
&\multicolumn{6}{c}{\bf Without External Information}\\\cline{2-8}
&SDE-Diffedit~\cite{nie2023theblessing}&	0.0659    &0.0211    &0.9032    &0.8513    &0.2705&4.2   \\
&NTI~\cite{mokady2023null}&0.0749&0.0197&0.8827&0.8206&0.2737&4.6\\	
&PNP~\cite{tumanyan2023plug}	&0.0970    &0.0244    &0.8781    &0.8252    &0.2739   &5.6	 \\
&HIVE~\cite{zhang2023hive}& 	0.1092    &0.0341    &0.8519    &0.7500    &0.2752 &6.6 \\				
&InstructP2P~\cite{brooks2023instructpix2pix}&	0.1122    &0.0371    &0.8524    &0.7428    &0.2764 &6.8 \\	
\cline{2-8}
&\multicolumn{6}{c}{\bf With External Information}\\\cline{2-8}
&GLIDE~\cite{nichol2022glide}&3.4973&115.8347&0.9487&0.9206&0.2249&6.0\\
&Blended Diffusion~\cite{Avrahami2023BlendedDiffusion}&3.5631&119.2813&0.9291&0.8644&0.2622&7.0\\
&Masactrl~\cite{cao2023masactrl}	&0.1570    &0.0501    &0.8120    &0.6797    &0.2768  &7.6 \\ 		
& MaSaFusion (DI-based)&0.0889    &0.0247    &0.9165    &0.8775    &0.2770& \textbf{3.2} \\
& MaSaFusion (NTI-based)&	0.0768    &0.0225    &0.9188    &0.8773    &0.2749  &3.4 \\
\hline			
\multirow{10}{*}{Multi-Turn} 
&\multicolumn{6}{c}{\bf Without External Information}\\\cline{2-8}
&SDE-Diffedit~\cite{nie2023theblessing}&	0.0831    &0.0279    &0.8813    &0.8023    &0.2719&3.2	\\
&NTI~\cite{mokady2023null}&0.1057&0.0335&0.8468& 0.7529&0.2710&5.2\\
&PNP~\cite{tumanyan2023plug}&0.1344   &0.0407    &0.8411    &0.7556    &0.2752&5.4\\
&HIVE~\cite{zhang2023hive}& 	0.1521    &0.0557    &0.8004    &0.6463    &0.2673	&7.2\\					
&InstructP2P~\cite{brooks2023instructpix2pix}&	0.1584    &0.0598    &0.7924    &0.6177    &0.2726&7.4	\\		
\cline{2-8}
&\multicolumn{6}{c}{\bf With External Information}\\\cline{2-8}
&GLIDE~\cite{nichol2022glide}&11.7487&1079.5997&0.9094&0.8494&0.2252&6.0\\
&Blended Diffusion~\cite{Avrahami2023BlendedDiffusion}& 14.5439&1510.2271& 0.8782&0.7690&0.2619&7.8\\
&Masactrl~\cite{cao2023masactrl}&	0.1966    &0.0735    &0.7856    &0.5958    &0.2777 & 7.8\\	
& MaSaFusion (DI-based)&   0.1103    &0.0340    &0.8961    &0.8425    &0.2807	& 3.0 \\
& MaSaFusion (NTI-based)&	0.0976    &0.0307    &0.9007    &0.8457    &0.2789 & \textbf{2.0}\\
\hline
\end{tabular}}}
\vspace{-0.2in}
\end{table}

    We compare our methods with training-based/free editing methods. For the first, we consider the standard methods HIVE \cite{zhang2023hive} and InstrctP2P \cite{brooks2023instructpix2pix}. For fair comparison, we do not fine-tune them on MagicBrush as in \cite{zhang2023magicbrush}. For training free methods, we mainly consider SDE-Diffedit \cite{couairon2022diffedit,nie2023theblessing}, P2P \cite{hertz2022prompt}, PnP \cite{tumanyan2023plug}, Masactrl \cite{cao2023masactrl}, GLIDE \cite{nichol2022glide}, and Blended Diffusion \cite{nie2023theblessing}. Here, the Masactrl, GLIDE, and Blended Diffusion use the external information as well, i.e., target sketch map for Masactrl, and editing region for the last two. These used conditions are exactly the ones used in our MaSaFusion.   
    
	\paragraph{Main Results.} We present some of the edited images in Figures \ref{fig:sve_fusion}\&\ref{fig:sve_fusion_multi_turn}. Moreover, we refer readers to more of them in Appendix \ref{app:More Results on SVE Task}. As can be seen, our method significantly improves the existing ones in terms of feature preserving and consistency to target prompt. 
    \par
    Besides, the quantitative comparisons with the baseline methods are in Table \ref{tbl:magicbrush}. For each method, we report its averaged rank (from 1-10) under each metric, and we have the following observations.
    \begin{enumerate}
        \item Owing to the diversity of consistency metrics, the representative metric should be the ``averaged rank'' of each method. By comparing them, we conclude that the proposed MaSaFusion indeed has the best performance over the other baseline methods.  
        \item The features in the edges of the editing region provided by the mask-informed methods GLIDE and Blended Diffusion are unsmooth and unnatural, which results in the large $L_{1}$ and $L_{2}$ distances. This originates from the directly copied out-of-mask features from the source image as clarified in Section \ref{sec:human annotation}.
        \item Compared with single-turn editing, implementing the existing methods of multi-turns generally preserves features at a poor level. We speculate this is because the unedited features from source images may gradually interrupted during the multi-turn editing, as its source images are outputs of former editing instead of the original ones.   
    \end{enumerate}


	\subsection{Ablation Study}\label{sec:ablation}
	In this subsection, we perform ablation studies on the additional T2I-Adapter's external conditions and the annotated editing regions. 
	\paragraph{Varying T2I-Adapter External Conditions.}
	In Figure \ref{fig:sve_fusion}, the external condition used in the T2I Adapter is a sketch map of the desired target image. In practice, the condition can be substituted by the ones that are suitable to the T2I Adapter e.g., pose or canny map as visualized in Figure \ref{fig:start_examples}. 
	
\begin{figure}[t!]
		\centering
		\vspace{-0.5in}
		\includegraphics[scale=0.25]{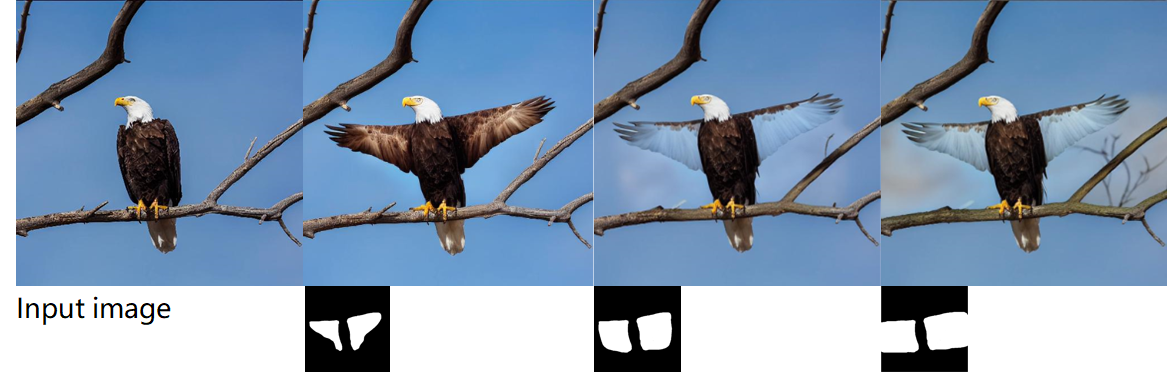}
		\caption{Variation of given editing region $A$ in our MaSaFusion. Obviously, the region should be somehow accurate, otherwise the editing is not satisfactory.}
		 \vspace{-0.2in}
		\label{fig:sve mask variation}
	\end{figure}

    \paragraph{Varying Editing Region.} 
	In MaSaFusion, the mask-informed region is predetermined, while it ideally should encompass the areas to be edited without overlapping with features that should remain unaltered. To check the sensitivity of MaSaFusion to the predetermined region, we vary the shape of the prior mask in Figure \ref{fig:sve mask variation}. As can be seen, an extremely large region can not retain the textures on the target image. We speculate such inappropriate region results in an undesired intermediate image in line 6 of Algorithm \ref{alg:masafusion editing}, which is borrowed to decide the target image's self-attention map. Thus, we suggest to invoke an appropriate editing region in MaSaFusion.        

	\section{Conclusion}
	In this paper, we improve the existing training-free diffusion-based text-to-image editing. Based on the failure cases of existing methods, we propose to generate intermediate hidden states with proper attention-map compared with the desired target image. During generating such intermediate image, we induce some external knowledge (i.e., the external conditions used in T2I Adapter). Then, these attention-maps are wisely borrowed in the following fusion process of generating target image, where the editing of target image is restricted in a prior given region.  
    \par
    We empirically verify our proposed method MaSaFusion on a benchmark T2I editing dataset. Compared to the existing methods, our MaSaFusion significantly improves the existing methods in T2I editing, both visually and quantitatively.

	%
	%
	\bibliographystyle{plain}
	\bibliography{reference}

\begin{thebibliography}{10}

\bibitem{Avrahami2023BlendedDiffusion}
Omri Avrahami, Dani Lischinski, and Ohad Fried.
\newblock Blended diffusion for text-driven editing of natural images.
\newblock In {\em Conference on Computer Vision and Pattern Recognition}, 2022.

\bibitem{brooks2023instructpix2pix}
Tim Brooks, Aleksander Holynski, and Alexei~A Efros.
\newblock Instructpix2pix: Learning to follow image editing instructions.
\newblock In {\em Conference on Computer Vision and Pattern Recognition}, 2023.

\bibitem{cao2023masactrl}
Mingdeng Cao, Xintao Wang, Zhongang Qi, Ying Shan, Xiaohu Qie, and Yinqiang
  Zheng.
\newblock Masactrl: Tuning-free mutual self-attention control for consistent
  image synthesis and editing.
\newblock Preprint arXiv:2304.08465, 2023.

\bibitem{caron2021emerging}
Mathilde Caron, Hugo Touvron, Ishan Misra, Herv{\'e} J{\'e}gou, Julien Mairal,
  Piotr Bojanowski, and Armand Joulin.
\newblock Emerging properties in self-supervised vision transformers.
\newblock In {\em International Conference on Computer Vision}, 2021.

\bibitem{couairon2022diffedit}
Guillaume Couairon, Jakob Verbeek, Holger Schwenk, and Matthieu Cord.
\newblock Diffedit: Diffusion-based semantic image editing with mask guidance.
\newblock In {\em International Conference on Learning Representations}, 2022.

\bibitem{esser2021taming}
Patrick Esser, Robin Rombach, and Bjorn Ommer.
\newblock Taming transformers for high-resolution image synthesis.
\newblock In {\em Conference on Computer Vision and Pattern Recognition}, 2021.

\bibitem{goodfellow2014generative}
Ian~J Goodfellow, Jean Pouget-Abadie, Mehdi Mirza, Bing Xu, David Warde-Farley,
  Sherjil Ozair, Aaron~C Courville, and Yoshua Bengio.
\newblock Generative adversarial nets.
\newblock In {\em Advances in Neural Information Processing Systems}, 2014.

\bibitem{gu2022vector}
Shuyang Gu, Dong Chen, Jianmin Bao, Fang Wen, Bo~Zhang, Dongdong Chen, Lu~Yuan,
  and Baining Guo.
\newblock Vector quantized diffusion model for text-to-image synthesis.
\newblock In {\em Conference on Computer Vision and Pattern Recognition}, 2022.

\bibitem{hertz2022prompt}
Amir Hertz, Ron Mokady, Jay Tenenbaum, Kfir Aberman, Yael Pritch, and Daniel
  Cohen-or.
\newblock Prompt-to-prompt image editing with cross-attention control.
\newblock In {\em International Conference on Learning Representations}, 2022.

\bibitem{ho2020denoising}
Jonathan Ho, Ajay Jain, and Pieter Abbeel.
\newblock Denoising diffusion probabilistic models.
\newblock In {\em Advances in Neural Information Processing Systems}, 2020.

\bibitem{ho2021classifier}
Jonathan Ho and Tim Salimans.
\newblock Classifier-free diffusion guidance.
\newblock In {\em NeurIPS 2021 Workshop on Deep Generative Models and
  Downstream Applications}, 2021.

\bibitem{ju2023direct}
Xuan Ju, Ailing Zeng, Yuxuan Bian, Shaoteng Liu, and Qiang Xu.
\newblock Direct inversion: Boosting diffusion-based editing with 3 lines of
  code.
\newblock Preprint arXiv:2304.04269, 2023.

\bibitem{kang2023scaling}
Minguk Kang, Jun-Yan Zhu, Richard Zhang, Jaesik Park, Eli Shechtman, Sylvain
  Paris, and Taesung Park.
\newblock Scaling up gans for text-to-image synthesis.
\newblock In {\em Conference on Computer Vision and Pattern Recognition}, 2023.

\bibitem{meng2021sdedit}
Chenlin Meng, Yutong He, Yang Song, Jiaming Song, Jiajun Wu, Jun-Yan Zhu, and
  Stefano Ermon.
\newblock Sdedit: Guided image synthesis and editing with stochastic
  differential equations.
\newblock In {\em International Conference on Learning Representations}, 2021.

\bibitem{mokady2023null}
Ron Mokady, Amir Hertz, Kfir Aberman, Yael Pritch, and Daniel Cohen-Or.
\newblock Null-text inversion for editing real images using guided diffusion
  models.
\newblock In {\em Conference on Computer Vision and Pattern Recognition}, 2023.

\bibitem{mou2023t2i}
Chong Mou, Xintao Wang, Liangbin Xie, Jian Zhang, Zhongang Qi, Ying Shan, and
  Xiaohu Qie.
\newblock T2i-adapter: Learning adapters to dig out more controllable ability
  for text-to-image diffusion models.
\newblock Preprint arXiv:2302.08453, 2023.

\bibitem{nichol2022glide}
Alexander~Quinn Nichol, Prafulla Dhariwal, Aditya Ramesh, Pranav Shyam, Pamela
  Mishkin, Bob Mcgrew, Ilya Sutskever, and Mark Chen.
\newblock Glide: Towards photorealistic image generation and editing with
  text-guided diffusion models.
\newblock In {\em International Conference on Machine Learning}, 2022.

\bibitem{nie2023theblessing}
Shen Nie, Hanzhong~Allan Guo, Cheng Lu, Yuhao Zhou, Chenyu Zheng, and Chongxuan
  Li.
\newblock The blessing of randomness: Sde beats ode in general diffusion-based
  image editing.
\newblock Preprint arXiv:2311.01410, 2023.

\bibitem{radford2021learning}
Alec Radford, Jong~Wook Kim, Chris Hallacy, Aditya Ramesh, Gabriel Goh,
  Sandhini Agarwal, Girish Sastry, Amanda Askell, Pamela Mishkin, Jack Clark,
  et~al.
\newblock Learning transferable visual models from natural language
  supervision.
\newblock In {\em International conference on machine learning}, 2021.

\bibitem{ramesh2022hierarchical}
Aditya Ramesh, Prafulla Dhariwal, Alex Nichol, Casey Chu, and Mark Chen.
\newblock Hierarchical text-conditional image generation with clip latents.
\newblock Preprint arXiv:2204.06125, 2022.

\bibitem{ramesh2021zero}
Aditya Ramesh, Mikhail Pavlov, Gabriel Goh, Scott Gray, Chelsea Voss, Alec
  Radford, Mark Chen, and Ilya Sutskever.
\newblock Zero-shot text-to-image generation.
\newblock In {\em International Conference on Machine Learning}, 2021.

\bibitem{rombach2022high}
Robin Rombach, Andreas Blattmann, Dominik Lorenz, Patrick Esser, and Bj{\"o}rn
  Ommer.
\newblock High-resolution image synthesis with latent diffusion models.
\newblock In {\em Conference on Computer Vision and Pattern Recognition}, 2022.

\bibitem{ronneberger2015u}
Olaf Ronneberger, Philipp Fischer, and Thomas Brox.
\newblock U-net: Convolutional networks for biomedical image segmentation.
\newblock In {\em Medical Image Computing and Computer-Assisted
  Intervention--MICCAI 2015: 18th International Conference, Munich, Germany,
  October 5-9, 2015, Proceedings, Part III 18}, pages 234--241. Springer, 2015.

\bibitem{saharia2022photorealistic}
Chitwan Saharia, William Chan, Saurabh Saxena, Lala Li, Jay Whang, Emily~L
  Denton, Kamyar Ghasemipour, Raphael Gontijo~Lopes, Burcu Karagol~Ayan, Tim
  Salimans, et~al.
\newblock Photorealistic text-to-image diffusion models with deep language
  understanding.
\newblock 2022.

\bibitem{sohl2015deep}
Jascha Sohl-Dickstein, Eric Weiss, Niru Maheswaranathan, and Surya Ganguli.
\newblock Deep unsupervised learning using nonequilibrium thermodynamics.
\newblock In {\em International Conference on Machine Learning}, 2015.

\bibitem{song2020denoising}
Jiaming Song, Chenlin Meng, and Stefano Ermon.
\newblock Denoising diffusion implicit models.
\newblock In {\em International Conference on Learning Representations}, 2022.

\bibitem{song2020score}
Yang Song, Jascha Sohl-Dickstein, Diederik~P Kingma, Abhishek Kumar, Stefano
  Ermon, and Ben Poole.
\newblock Score-based generative modeling through stochastic differential
  equations.
\newblock In {\em International Conference on Learning Representations}, 2020.

\bibitem{su2021pixel}
Zhuo Su, Wenzhe Liu, Zitong Yu, Dewen Hu, Qing Liao, Qi~Tian, Matti
  Pietik{\"a}inen, and Li~Liu.
\newblock Pixel difference networks for efficient edge detection.
\newblock In {\em International Conference on Computer Vision}, 2021.

\bibitem{tang2023daam}
Raphael Tang, Akshat Pandey, Zhiying Jiang, Gefei Yang, Karun Kumar, Jimmy Lin,
  and Ferhan Ture.
\newblock What the daam: Interpreting stable diffusion using cross attention.
\newblock 2023.

\bibitem{tumanyan2023plug}
Narek Tumanyan, Michal Geyer, Shai Bagon, and Tali Dekel.
\newblock Plug-and-play diffusion features for text-driven image-to-image
  translation.
\newblock In {\em Conference on Computer Vision and Pattern Recognition}, 2023.

\bibitem{vaswani2017attention}
Ashish Vaswani, Noam Shazeer, Niki Parmar, Jakob Uszkoreit, Llion Jones,
  Aidan~N Gomez, {\L}ukasz Kaiser, and Illia Polosukhin.
\newblock Attention is all you need.
\newblock {\em Advances in Neural Information Processing Systems}, 2017.

\bibitem{xue2023sa}
Shuchen Xue, Mingyang Yi, Weijian Luo, Shifeng Zhang, Jiacheng Sun, Zhenguo Li,
  and Zhi-Ming Ma.
\newblock Sa-solver: Stochastic adams solver for fast sampling of diffusion
  models.
\newblock Preprint arXiv:2309.05019, 2023.

\bibitem{yang2023paint}
Binxin Yang, Shuyang Gu, Bo~Zhang, Ting Zhang, Xuejin Chen, Xiaoyan Sun, Dong
  Chen, and Fang Wen.
\newblock Paint by example: Exemplar-based image editing with diffusion models.
\newblock In {\em Conference on Computer Vision and Pattern Recognition}, 2023.

\bibitem{yi2023generalization}
Mingyang Yi, Jiacheng Sun, and Zhenguo Li.
\newblock On the generalization of diffusion model.
\newblock Preprint arXiv:2305.14712, 2023.

\bibitem{yu2022scaling}
Jiahui Yu, Yuanzhong Xu, Jing~Yu Koh, Thang Luong, Gunjan Baid, Zirui Wang,
  Vijay Vasudevan, Alexander Ku, Yinfei Yang, Burcu~Karagol Ayan, et~al.
\newblock Scaling autoregressive models for content-rich text-to-image
  generation.
\newblock {\em Transactions on Machine Learning Research}, 2022.

\bibitem{zhang2021cross}
Han Zhang, Jing~Yu Koh, Jason Baldridge, Honglak Lee, and Yinfei Yang.
\newblock Cross-modal contrastive learning for text-to-image generation.
\newblock In {\em Conference on Computer Vision and Pattern Recognition}, 2021.

\bibitem{zhang2017stackgan}
Han Zhang, Tao Xu, Hongsheng Li, Shaoting Zhang, Xiaogang Wang, Xiaolei Huang,
  and Dimitris~N Metaxas.
\newblock Stackgan: Text to photo-realistic image synthesis with stacked
  generative adversarial networks.
\newblock In {\em International Conference on Computer Vision}, 2017.

\bibitem{zhang2018stackgan++}
Han Zhang, Tao Xu, Hongsheng Li, Shaoting Zhang, Xiaogang Wang, Xiaolei Huang,
  and Dimitris~N Metaxas.
\newblock Stackgan++: Realistic image synthesis with stacked generative
  adversarial networks.
\newblock {\em IEEE transactions on pattern analysis and machine intelligence},
  41(8):1947--1962, 2018.

\bibitem{zhang2023magicbrush}
Kai Zhang, Lingbo Mo, Wenhu Chen, Huan Sun, and Yu~Su.
\newblock Magicbrush: {A} manually annotated dataset for instruction-guided
  image editing.
\newblock In {\em Conference on Neural Information Processing Systems}, 2023.

\bibitem{zhang2023adding}
Lvmin Zhang, Anyi Rao, and Maneesh Agrawala.
\newblock Adding conditional control to text-to-image diffusion models.
\newblock In {\em International Conference on Computer Vision}, 2023.

\bibitem{zhang2023hive}
Shu Zhang, Xinyi Yang, Yihao Feng, Can Qin, Chia-Chih Chen, Ning Yu, Zeyuan
  Chen, Huan Wang, Silvio Savarese, Stefano Ermon, et~al.
\newblock Hive: Harnessing human feedback for instructional visual editing.
\newblock Preprint arXiv:2303.09618, 2023.

\bibitem{zhang2023sine}
Zhixing Zhang, Ligong Han, Arnab Ghosh, Dimitris~N Metaxas, and Jian Ren.
\newblock Sine: Single image editing with text-to-image diffusion models.
\newblock In {\em Conference on Computer Vision and Pattern Recognition}, 2023.

\end{thebibliography}
	
	\appendix
	\section{Discussion to Inversion Techniques}\label{app:null-text}
\begin{algorithm}[t!]
	\caption{Null-text Inversion}
	\label{alg:null-text inversion}
	\textbf{Input:} source prompt $\cS$, source image $\bx^{\cS}_{0}$, diffusion model $\beps_{\btheta}$, textual model $\tau$, null-text optimization steps $N$.   
	\begin{algorithmic}[1]
		\STATE  {Guidance scale $w = 1$,}
		\STATE  {Conduct DDIM inversion \eqref{eq:DDIM inversion} to get intermediate states $\{\bx_{t}^{\cS}\}_{t=1}^{T}$;}
		\STATE  {Guidance scale $w = 7.5$;}
		\STATE  {Initialize $\hat{\bx}_{T}^{\cS} = \bx_{T}^{\cS}$, $\emptyset_{T} = \tau("")$;}
		\FOR    {$t = T, \cdots ,1$}
		\FOR   {$j = 0, \cdots, N - 1$}
		\STATE  {$\emptyset_{t} = \emptyset_{t} - \eta\nabla_{\emptyset}\|\bx_{t - 1}^{\cS} - \mathrm{DDIM}(\beps_{\btheta}, \hat{\bx}_{t}^{\cS}, \emptyset_{t}, \tau(\cS))\|$;}
		\ENDFOR 
		\STATE  {$\hat{\bx}_{t - 1}^{\cS} = \mathrm{DDIM}(\hat{\bx}_{t}^{\cS}, \emptyset_{t}, \cS)$, $\emptyset_{t - 1} = \emptyset_{t}$;}
		\ENDFOR
		\RETURN {Initialize noise $\bx^{\cS}_{T}$, null-text embedding set $\{\emptyset_{t}\}$.}
	\end{algorithmic}
\end{algorithm}

In this section, we present a brief introduction to the technique of null-text inversion \cite{mokady2023null}, which is a backbone inversion technique of our method. It has been found in \cite{mokady2023null}, the standard DDIM inversion \eqref{eq:DDIM inversion} with source prompt injected into $\beps_{\btheta}(t, \bx_{t}^{\cS}, \cC^{\cS}, \emptyset)$ can not reconstruct the source image. Thus they propose to learn a series of $\{\emptyset_{t}\}_{t = 1}^{T}$ by iteratively minimizing 
\begin{equation}\label{eq:nti objective}
	\small
	\min_{\emptyset_{t}}\left\|\bx_{t - 1}^{\cS} - \mathrm{DDIM}(\beps_{\btheta}, \hat{\bx}_{t}^{\cS}, \cC, \emptyset_{t})\right\|^{2},
\end{equation} 
where $\bx_{t}^{\cS}$ is obtained by DDIM inversion \eqref{eq:DDIM inversion}, starting from source image $\bx_{0}^{\cS}$, and $\hat{\bx}_{t}^{\cS}$ is obtained by DDIM \eqref{eq:DDIM} from $\hat{\bx}_{t + 1}^{\cS}$ with $\hat{\bx}_{T}^{\cS} = \bx_{T}^{\cS}$ and $\emptyset = \emptyset_{t}$ in \eqref{eq:noise prediction}. Then, plugging these $\{\emptyset_{t}\}$ into \eqref{eq:DDIM} from $\bx_{T}^{\cS}$ reconstructs the source image. The complete NTI is summarized in Algorithm \ref{alg:null-text inversion} \cite{mokady2023null}. 

The null-text inversion incorporates the information from the source prompt into the learned initial state $\bx_{0}^{\cT}$ and $\{\emptyset_{t}\}$ so that it is friendly to preserve features when generating target image. In this paper, the hyperparameters of null-text inversion are adopted from \cite{mokady2023null} with $N = 10, T = 50, \eta = 2e-4$. We refer readers for more details of this method in \cite{mokady2023null}.   

\begin{algorithm}[t!]
	\caption{Direct Inversion}
	\label{alg:direct inversion}
	\textbf{Input:} source prompt $\cS$, source image $\bx^{\cS}_{0}$, diffusion model $\beps_{\btheta}$, textual model $\tau$, null-text optimization steps $N$.   
	\begin{algorithmic}[1]
		\STATE  {Guidance scale $w = 1$,}
		\STATE  {Conduct DDIM inversion \eqref{eq:DDIM inversion} to get intermediate states $\{\bx_{t}^{\cS}\}_{t=1}^{T}$;}
		\STATE  {Guidance scale $w = 7.5$;}
		\STATE  {Initialize $\hat{\bx}_{T}^{\cS} = \bx_{T}^{\cS}$, $\emptyset_{T} = \tau("")$;}
		\FOR    {$t = T, \cdots ,1$}
		\STATE  {Implementing $\mathrm{DDIM}(\beps_{\btheta}, \bx_{t}^{\cS}, \emptyset, \tau(\cS))$;}
		\STATE  {Obtain $\{(Q^{\cS}, K^{\cS}, V^{\cS})\}$ for each $t$ to use in the followed fusion process;}
		\ENDFOR
		\RETURN {Initialize noise $\bx^{\cS}_{T}$.}
	\end{algorithmic}
\end{algorithm}

In NTI Algorithm \ref{alg:null-text inversion}, obtaining the null-conditions $\{\emptyset_{t}\}$ requires an extra inner loop of optimization. The inner is used to make $\hat{\bx}_{t}^{\cS}$ approximates $\bx_{t}^{\cS}$ so that fits the text prompt condition with $w=7.5$ in generating target image. However, the extra inner loop increases the computational complexity, which somehow restricts the editing technique in practice. Recently, another inversion technique Direct Inversion \cite{ju2023direct} was proposed to accelerate NTI. It improves the NTI by canceling the inner loop in lines 6-8 of Algorithm \ref{alg:null-text inversion} to accelerate the editing process. The idea in \cite{ju2023direct} is simple, instead of optimizing $\{\emptyset_{t}\}$ to lead $\hat{\bx}_{t}^{\cS}$ approximates $\bx_{t}^{\cS}$, they directly make $\hat{\bx}_{t}^{\cS} = \bx_{t}^{\cS}$ and feed them into the DDIM process for $w=7.5$ to obtain intermediate hidden states  $\{(Q^{\cS}, K^{\cS}, V^{\cS})\}$.    
Notably, combining MaSaFusion with DI slightly drops its performance, compared with conducting it under NTI, as in Table \ref{tbl:magicbrush}. However, DI decreases the whole editing process from 184.2s to 43.9s (evaluated on a workstation with one NVIDIA V100). Thus, we prefer to use DI as our backbone inversion technique in practice.  

\section{Shape Invariant Editing}\label{sec:SIE}
\begin{figure}[t!]
	\centering
	\includegraphics[scale=0.3]{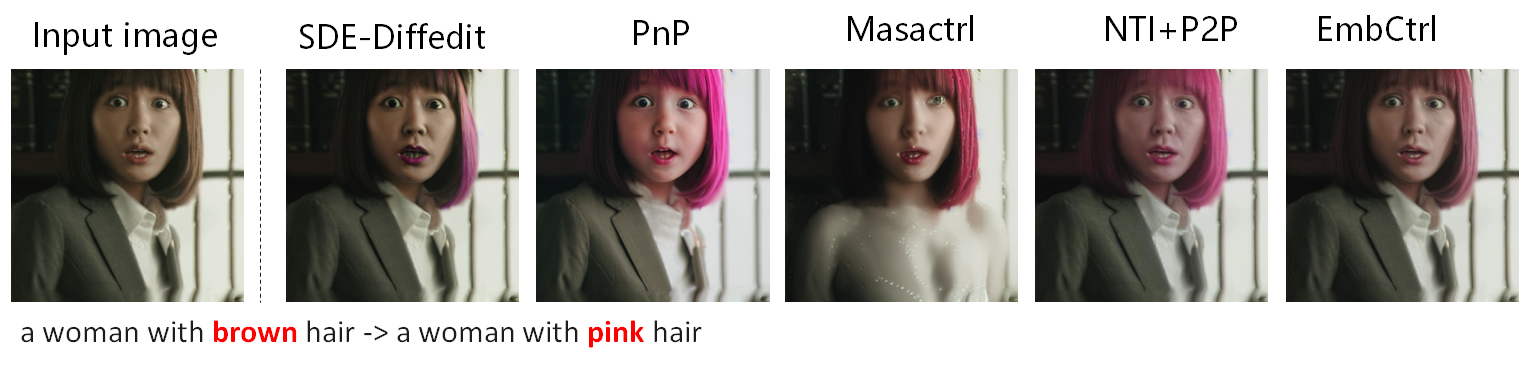}
	\caption{The results of existing methods under shape invariant editing task.}
	\label{fig:reconstruction}
\end{figure}
In this section, we explore Text-to-Image (T2I) editing tasks where the shapes of the features being edited remain largely unchanged. The typical scenarios include background or color transformations. As we have clarified in Section \ref{sec:interference When Doing SVE}, the existing methods (e.g., P2P) have better performance under these tasks than the other shape variant editing tasks. However, they can be further improved without the external knowledge as in MaSaFusion. 
\subsection{Control Unrelated Textual Embedding}\label{sec:Unrelated Textual Embedding Deteriorate Preserving}
\begin{figure}[t!]
	\centering
	\includegraphics[scale=0.45]{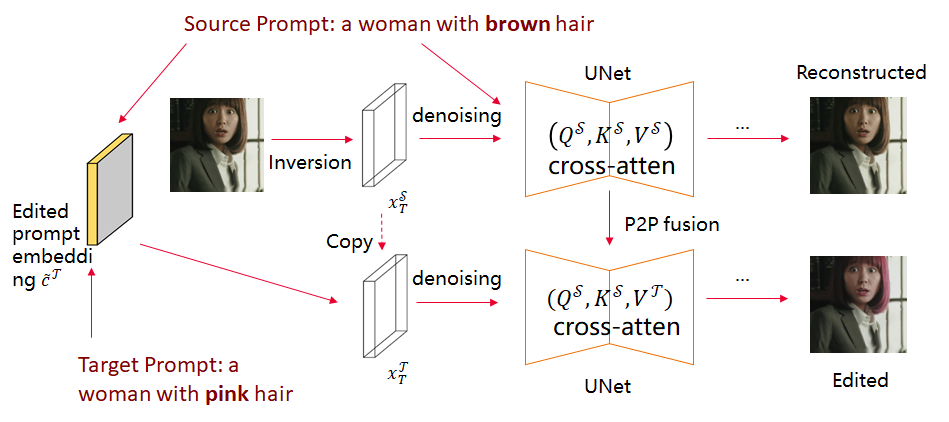}
	\caption{The visualization of our method for shape invariant editing, i.e., EmbCtrl Algorithm \ref{alg:sie editing}.}
	\label{fig:sie method}
\end{figure}
\begin{algorithm}[t!]
	\caption{EmbCtrl.}
	\label{alg:sie editing}
	\textbf{Input:} Source prompt and image ($\cS$, $\bx^{\cS}_{0}$), target prompt $\cT$, diffusion model $\beps_{\btheta}$.  
	\begin{algorithmic}[1]
		\STATE  {\emph{\textbf{Inversion Step}}}
		\STATE  {Inversion Algorithm \ref{alg:null-text inversion} or \ref{alg:direct inversion} on $(\cS, \bx^{\cS}_{0})$ to get $\{\emptyset_{t}\}$ (NTI) and $\bx^{\cS}_{T}$;} 
		\STATE  {Create new target prompt embedding $\tilde{\cC}^{\cT}$ as in \eqref{eq:new prompt};}
		\STATE  {\emph{\textbf{Generating Target Image}}}
		\STATE  {Initialize $\bx^{\cT}_{T} = \bx_{T}^{\cS}$;}
		\FOR    {$t = T, \cdots ,1$}
		\STATE  {$\bx_{t - 1}^{\cT}=\mathrm{DDIM}(\beps_{\btheta}, \bx^{\cT}_{t}, \tilde{\cC}^{\cT}, \emptyset)$ (process of \eqref{eq:DDIM inversion});} 
		\ENDFOR
		\RETURN {Target image $\bx^{\cT}_{0}$.}
	\end{algorithmic}
\end{algorithm}
In the existing methods, though the edited target image roughly preserves the characteristics of the source image, the overall performance is far from satisfactory, e.g., the characteristic of humanity face varies in Figure \ref{fig:reconstruction}. In the rest of this subsection, we aim to improve the null-text-inversion, and fix the issues as in Figure \ref{fig:reconstruction}. 
\par
Due to \eqref{eq:cross-attent}, the textual prompt affects the generated image latent by cross-attention structure. For the intermediate hidden states $\phi(\bx_{t})\in \bbR^{N\times d}$ (flattened), where $N$ is the number of pixels and $d$ is the dimension of pixel-wise embedding, the $i$-pixel of cross-attention's output is $\sum_{j=1}^{d_{\cC}}P_{ij}V_{j}$, 
where $P = \mathrm{Softmax}(QK^\top/\sqrt{d})$, $V=W_{V}\cC$. Thus, the representation of each pixel after the cross-attention layer is indeed a weighted sum of $V$ which is a linear transformation of textual embedding $\cC$ obtained by CLIP \cite{radford2021learning} model. Ideally, the pixels in a specific feature of the generated image are only decided by the semantically related textual prompt. Then, the features can be appropriately edited by substituting tokens in the source prompt.
\par
\begin{figure}[t!]
	\centering
	\includegraphics[scale=0.5]{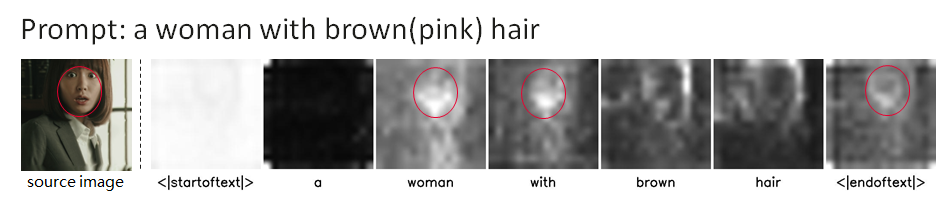}
	\caption{The visualization cross attention map of image with given source prompt.}
	\label{fig:attention_mask}
\end{figure}

Unfortunately, this can not happen in practice. In Figure \ref{fig:attention_mask}, by following \cite{tang2023daam}, we visualize the cross-attention map of softmax for each word in prompt $\cC$ ``a woman with brown hair''. It is worthy to note that $\cC$ is padded by ``$\mathrm{<|endoftext|>}$'' into a length of 77 as in \cite{rombach2022high}. As can be seen, the unedited feature, \footnote{Interestingly, though the image is encoded latent, the token can semantically relate to the corresponding image. Similar results are also observed in \cite{tang2023daam}.} e.g., the human face is decided by the other unchanged tokens (``woman'', ``with'', ``$\mathrm{<|endoftext|>}$''). This dependence causes the failure of null-text-inversion. Because when altering the source textual prompt ``woman with brown hair'' into the target textual prompt ``woman with pink hair'', the embedding of tokens after ``pink'' are all changed due to the auto-regressive embedding mechanism of CLIP model \cite{radford2021learning}. Then, the entirely changed target textual prompt affects the overall features in the target image.

\begin{figure}[t!]
	\centering
	\includegraphics[scale=0.3]{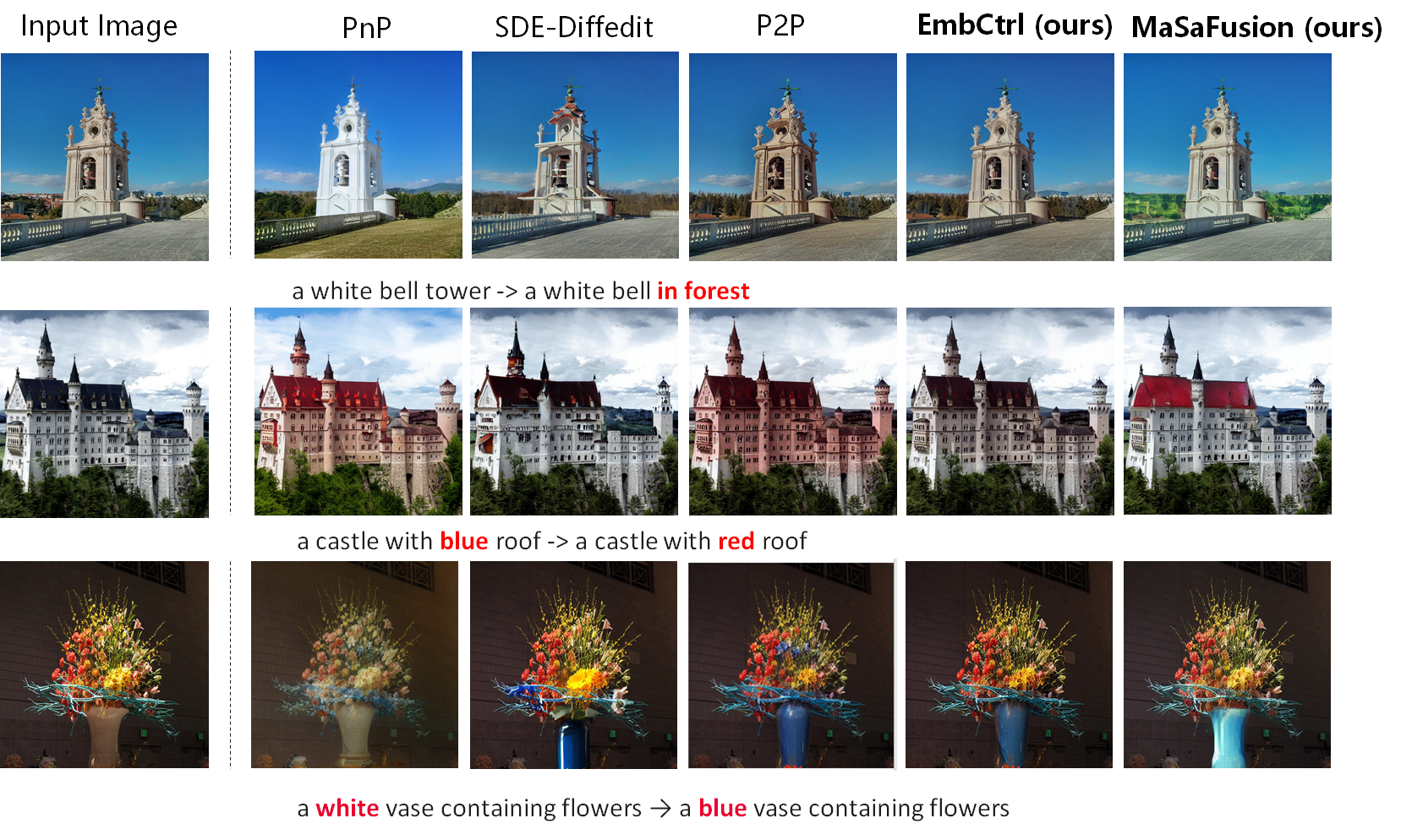}
	\caption{The comparison over editing methods on shape invariant editing task. The compared baseline methods are mainly tailored to such editing task.}
	\label{fig:sie_results}
\end{figure}
\par
To obviate this, a straightforward way of improving null-text-inversion is keeping the token embedding in the target prompt for unchanged tokens. That says, given textual embedding $\cC^{\cS} = \{\tau(\cS)_{1}, \cdots, \tau(\cS)_{77}\}$ ($\tau(\cdot)$ is CLIP model) of source prompt $\cS$, we use the target prompt embedding   
\begin{equation}\label{eq:new prompt}
	\small
	\tilde{\cC}^{\cT} = \{\tau(\cS)_{1}, \cdots, \tau(\cT)_{i}, \cdots, \tau(\cS)_{77}\}, 
\end{equation}   
instead of $\cC^{\cT} = \tau(\cT)$, where $\tau(\cT) = \cC^{\cT}$ is the textual embedding of target prompt $\cT$ differs from source prompt $\cS$ in the $i$-th token ($i$ can be larger than 1). The pipeline of our method ``Embedding Control'' (EmbCtrl) to tackle shape invariant editing task is in Algorithm \ref{alg:sie editing} where $\mathrm{DDIM}(\beps_{\btheta}, \bx^{\cT}_{t}, \tilde{\cC}^{\cT}, \emptyset)$ is defined as in \eqref{eq:nti objective}. 
\par
The edited images of our method are in Figure \ref{fig:sie_results}. As can be seen, compared with the existing methods, our method maximally preserves the features of source images that are undesired to be edited. Besides that, the proposed EmbCtrl does not require extra external conditions as in the proposed MaSaFusion. 
\par
However, the proposed EmbCtrl can not be generally applied as MaSaFusion (for both shape invariant/variant editing). For those editing with variant objects' shapes, the EmbCtrl will fail similarly to the other simple fusion-based methods (e.g., P2P). 




\section{MaSaFusion without Intermediate Images}\label{app:SVE without Intermediate Images}
As we have clarified in Section \ref{sec:interference When Doing SVE}, generating an intermediate image with the correct attention map is critical to the final fusion process. In this section, we verify this clarification. Instead of involving an external signal (e.g., sketch) to capture intermediate hidden states by T2I adapter as in Algorithm \ref{alg:masafusion editing}, we develop another pipeline to conduct text-to-image editing in the sequel. It is worth noting that MaSaFusion leverages human annotations in two-fold, i.e., external signal and editing region. We try to use none of them at first, and we will observe that without any of them, the editing is unstable.  
\par
Firstly, our method is as follows: for the procedure of DDIM in line 7 of Algorithm \ref{alg:sie editing}, we modify the noise prediction $\beps_{\btheta}(t, \bx_{t}, \cC^{\cT}, \emptyset_{t})$ as follow. For the self-attention modules in the network, under specific $t$, we substitute the original output $\mathrm{Attention}_{SA}(Q^{\cT}, K^{\cT}, V^{\cT})$ with $\mathrm{Attention}_{SA}(Q^{\cT}, K^{\cS}, V^{\cS})$ for the corresponded layers. \footnote{According to our empirical observations, this works better than the other blocks, i.e., cross-attention and residual blocks.} Besides that, instead of $\emptyset_{t}$, we should use empty embedding $\emptyset$ when generating target image. The substituted noise prediction is denoted as
\begin{equation}\label{eq:bar eps}
	\small
	\hat{\beps}_{\btheta}^{\mathrm{w/o\ t2i}}(t, \bx_{t}, \cC^{\cT}, \cC^{\cS}, \emptyset)
\end{equation}
in the sequel to clarify the notation. The rationale of substituting $(Q^{\cT}, K^{\cT}, V^{\cT})$ with $(Q^{\cT}, K^{\cS}, V^{\cS})$ in self-attention module is intuitive because 1): The self-attention decides output by the image itself instead of target prompt as in cross-attention module (which generates unrelated image); 2): We want the generated features to be the composition of source image but in the way of target prompt, so that from $(Q^{\cT}, K^{\cT}, V^{\cT})$ to $(Q^{\cT}, K^{\cS}, V^{\cS})$.    
\par
\begin{figure*}[t!]
	\centering
	\includegraphics[scale=0.3]{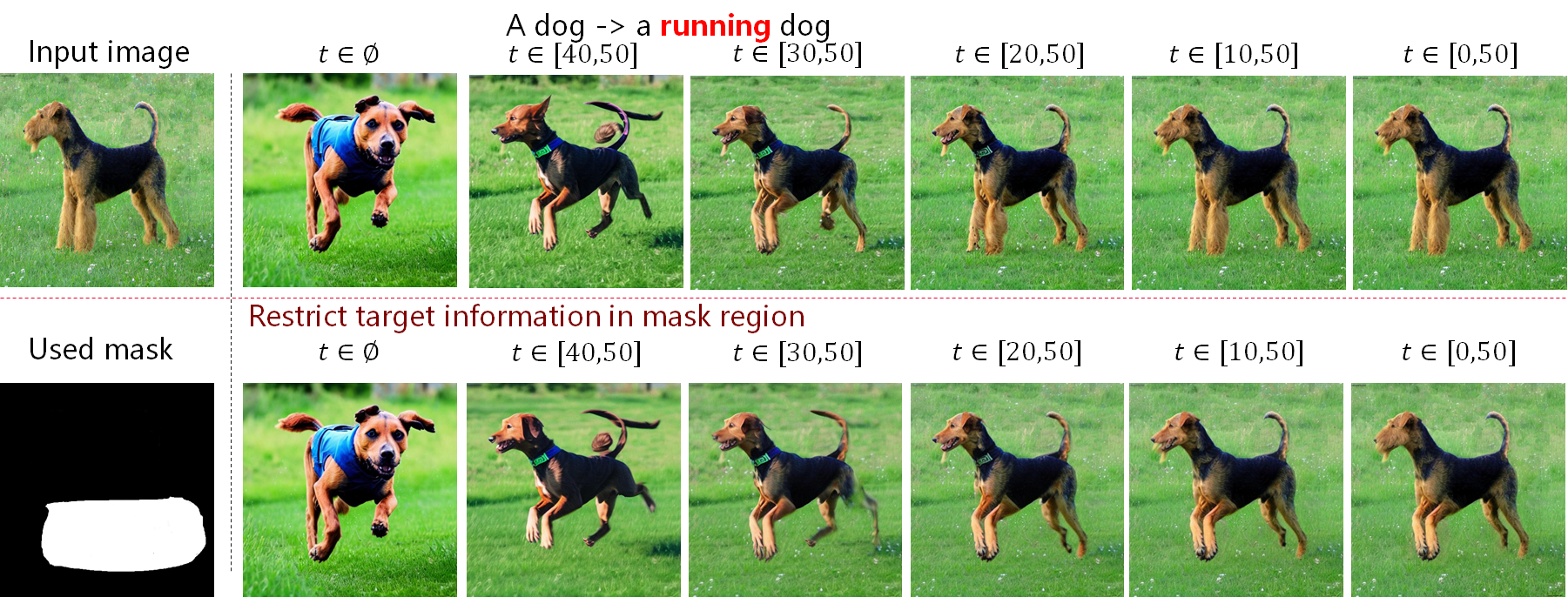}
	\caption{The edited images of injecting information from source prompt as in \eqref{eq:bar eps} over different time interval of DDIM \eqref{eq:DDIM}. In the first (resp. second) row, the interval $t\in[a, b]$ means substituting $\beps_{\btheta}$ in DDIM \eqref{eq:DDIM} with noise prediction defined in \eqref{eq:bar eps} (resp. \eqref{eq:mask eps}) when $t\in [a, b]$. Please note that in the row of images, the target information are restricted in the mask region.}
	\label{fig:sve_fusion wo t2i}
\end{figure*}
\par
Let us check the results without any human annotations. We vary the number of denoising steps that inject information from the source prompt in Figure \ref{fig:sve_fusion wo t2i}. In the first row of Figure \ref{fig:sve_fusion wo t2i}, the strength of artificial fusion is increased from the left to right. As can be seen, without the fusion operation, the generated image is consistent with the target prompt, while the characteristics are slightly related to the source image. On the other hand, increasing the strength of fusion rends the generated image close to the source one until degeneration happens. These observations inform that wisely modify the self-attention block of noise prediction $\beps_{\btheta}(t, \bx_{t}, \cC^{\cT}, \emptyset)$ in DDIM \eqref{eq:DDIM} potentially works, e.g., conducting fusion on the first 20 steps of DDIM \eqref{eq:DDIM} in Figure \ref{fig:sve_fusion wo t2i}. However, this setting is not necessarily generalized to the other images.   
\subsection{Injecting Prior Knowledge Editing Region}
\begin{algorithm}[t!]
	\caption{MaSaFusion without T2I Adapter.}
	\label{alg:sve editing}
	\textbf{Input:} source prompt $\cS$, source image $\bx^{\cS}_{0}$, target prompt $\cT$, diffusion model $\beps_{\btheta}$, annotation mask, null-text embedding $\emptyset=\tau("")$.   
	\begin{algorithmic}[1]
		\STATE  {\emph{\textbf{Inversion Step}}}
		\STATE  {Inversion Algorithm \eqref{alg:null-text inversion} or \ref{alg:direct inversion} on $(\cS, \bx^{\cS}_{0})$ to get $\{\emptyset_{t}\}$ (NTI) and $\bx^{\cS}_{T}$;} 
		\STATE  {DDIM \eqref{eq:DDIM} with $\bx^{\cS}_{T}$, $\{\emptyset_{t}\}$ and get intermediate hidden states $\{Q^{\cS}, K^{\cS}, V^{\cS}\}$ for each $t$;}
		\STATE  {\emph{\textbf{Generating Target Image}}}
		\STATE  {Initialize $\bx^{\cT}_{T} = \bx_{T}^{\cS}$;}
		\STATE {\textbf{\emph{Fuse cached intermediate hidden states $\{Q^{\cS}, K^{\cS}, V^{\cS}\}$}}}
		\STATE  {$\bx_{t - 1}^{\cT}=\mathrm{DDIM}(\beps_{\btheta}, \bx^{\cT}_{t}, \cC^{\cT}, \emptyset)$ \eqref{eq:mask eps};}
		\STATE {\textbf{\emph{Fuse cached intermediate hidden states $\{Q^{\cS}, K^{\cS}, V^{\cS}\}$}}
			\newline 
			DDIM \eqref{eq:DDIM} with $\bepsilon_{\btheta}$ in \eqref{eq:eps masafusion without t2i}, by invoking annotation region $A$;}
		\RETURN {Target image $\bx^{\cT}_{0}$.}
	\end{algorithmic}
\end{algorithm}
In this subsection, we promote the method in above by involving editing region. To do so, we revise the noise-prediction as 
\begin{equation}\label{eq:eps masafusion without t2i}
	\small
	\beps_{\btheta}(i) =
	\begin{cases}
		&\bar{\beps}_{\btheta}(t, \bx_{t}^{\cT}, \cC^{\cS}, \emptyset), \qquad \qquad i\notin A; \\
		&\hat{\beps}_{\btheta}^{\mathrm{w/o\ t2i}}(t, \bx_{t}, \cC^{\cT}, \cC^{\cS}, \emptyset), \quad i\in A,
	\end{cases}	
\end{equation}
\begin{figure}[h]
	\centering
	\includegraphics[scale=0.5]{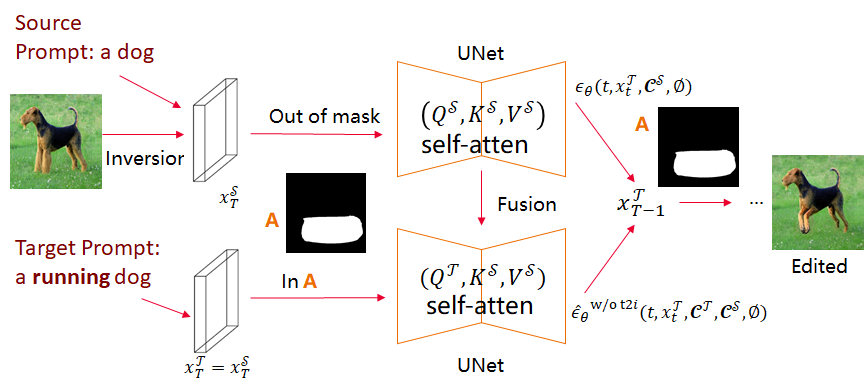}
	\caption{The visualization of our method without T2I adapter \cite{mou2023t2i} for shape variant editing, i.e., MaSaFusion without T2I adapter Algorithm \ref{alg:sve editing}.}
	\label{fig:sve method}
\end{figure}
to restrict the potential interfere of generating new features and preserving source image in the prior editing region, where $\bar{\beps}_{\btheta}(t, \bx_{t}^{\cT}, \cC^{\cS}, \emptyset_{t})$ is defined in \eqref{eq:mask eps}. By leveraging the annotation region, the interference is greatly mitigated since the generated new features are restricted in a small region. Besides that, with this mask, injecting the information from the source prompt can be conducted over the whole DDIM process as in Figure \ref{fig:sve_fusion wo t2i}, so obviating the collapse into the source image. The pipeline of this method is summarized in Figure \ref{fig:sve method} and Algorithm \ref{alg:sve editing}. 
\par
However, though the Algorithm \ref{alg:sve editing} works well in some cases, we observe the method is unstable in practice. For example, in Figure \ref{fig:fusion process}, the generated eagle is unrealistic. Comparing the existing results, the generated dog for $t\in\emptyset$ and source image in Figure \ref{fig:sve_fusion wo t2i} accidentally has legs in the same prior editing region, so that the fusion between them is relatively simple. However, for the eagle in Figure \ref{fig:fusion process}, the edited characteristics in these two images are not in the same prior editing region. Thus, the mismatch results in an inappropriate fusion process. 
\par
To obviate this, we should guarantee the generated image to be fused has edited characteristics in the prior given editing region. One way to doing this is leveraging the T2I Adapter as in our MaSaFusion \ref{alg:masafusion editing}.

\section{More Results of MaSaFusion}\label{app:More Results on SVE Task}
\begin{figure*}[t]
	 	\centering
	 	\includegraphics[scale=0.38]{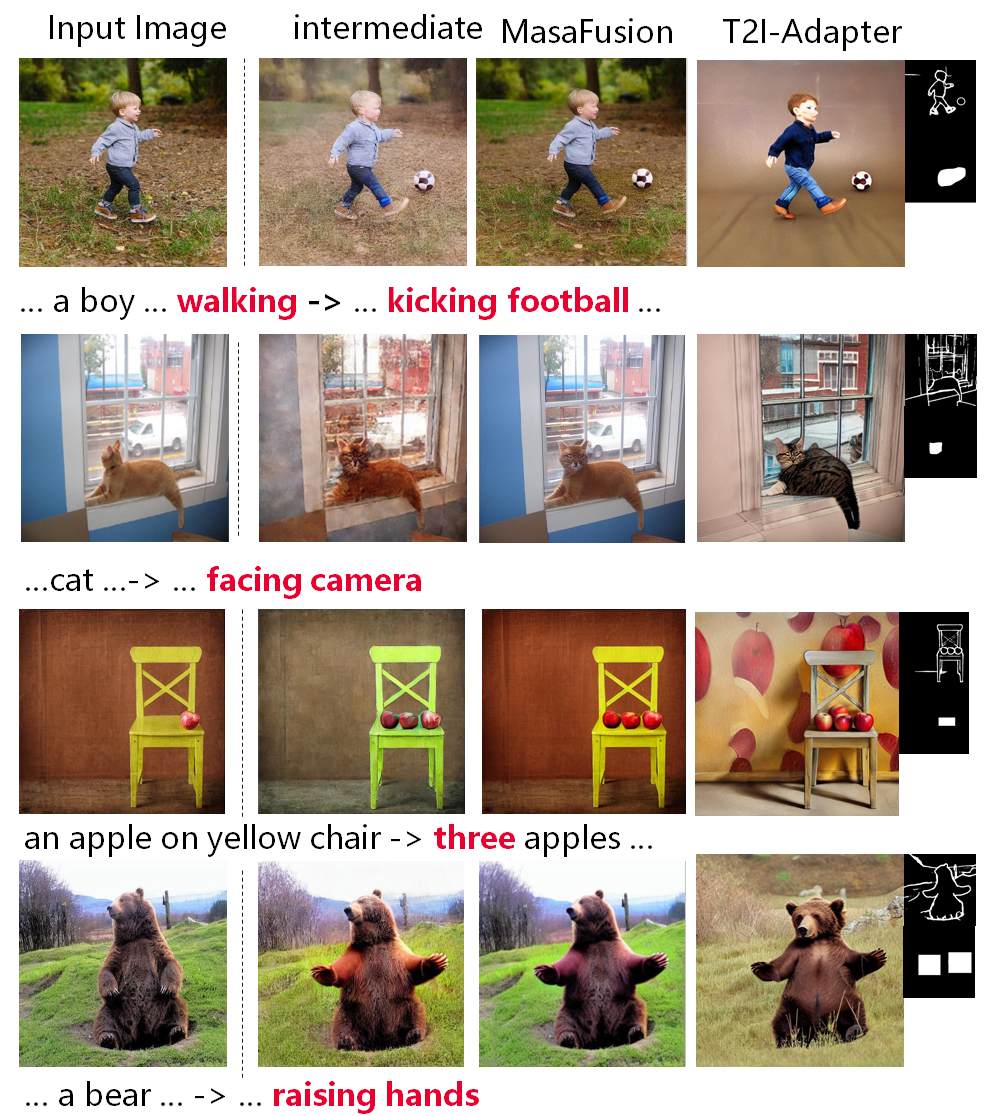}
	 	\caption{The comparison of source image, intermediate generated image, generated target image, and T2I Adapter generated image.}
	 	\label{fig:sve_intermediate}
\end{figure*}

In this section, we show more results and discussion of our MaSaFusion \ref{alg:masafusion editing}. Firstly, we present some intermediate results on our method, i.e., the generated image used to be fused in line 6 of Algorithm \ref{alg:masafusion editing} and the target image directly obtained by T2I Adapter. The results are in Figure \ref{fig:sve_intermediate}. As can be seen, though the intermediate images and images generated by the T2I Adapter are consistent with the target prompt, some features are not preserved as we hope. This explains our claims about them in Section \ref{sec:human annotation}. Besides that, one can notice that the generated target image has an inconsistent facial identity for images in the first row of Figure \ref{fig:sve_intermediate}. This originates from the limitation of Stable Diffusion in some cases of generation e.g., human face or hand. 
\par
\begin{figure}[t]
 	\centering
	 	\includegraphics[scale=1.0]{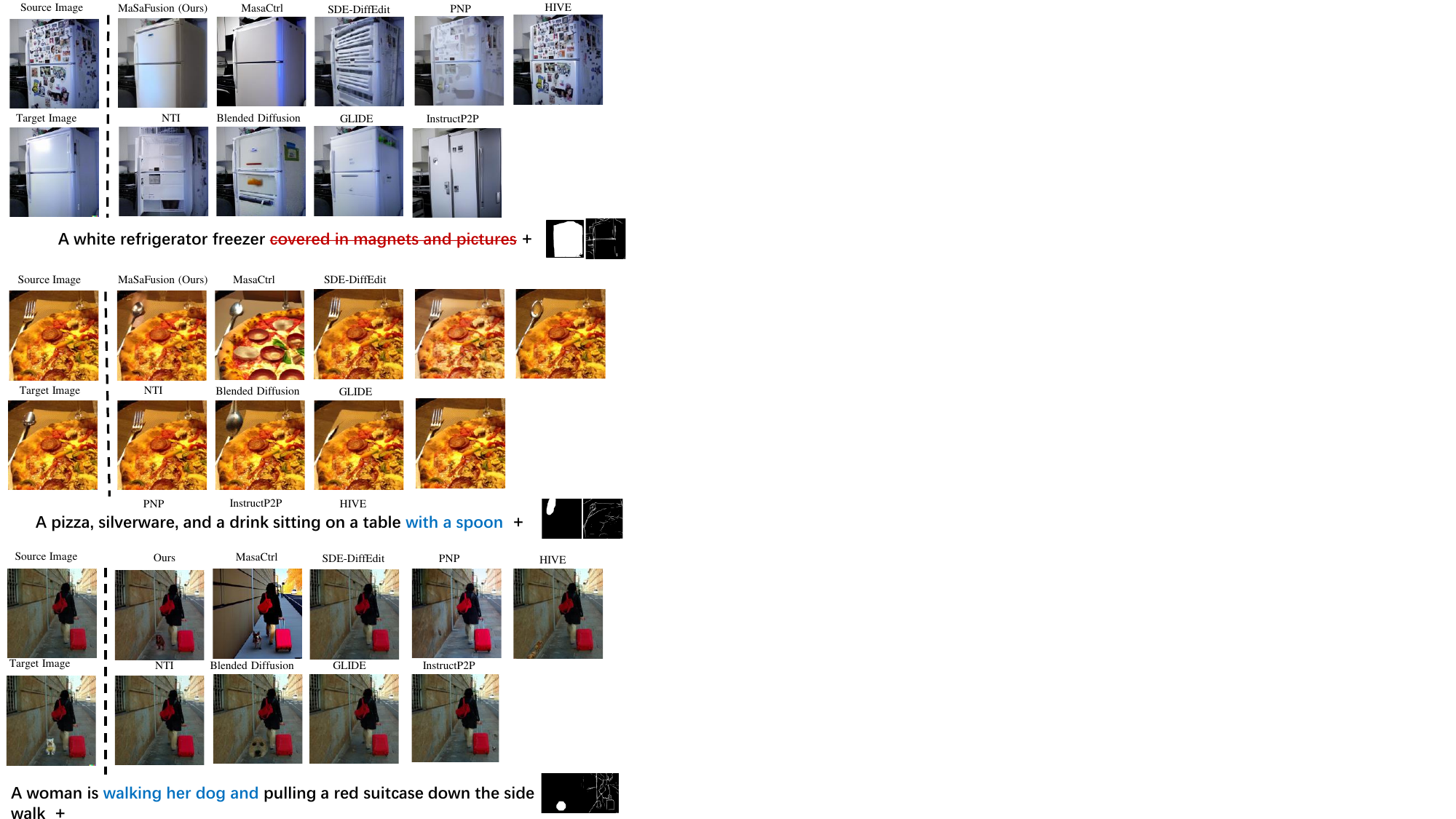}
	 	\caption{The comparison of our MaSaFusion with existing methods.}
	 	\label{fig:sve_more_masafusion}
\end{figure}

Moreover, we present more edited images of our and other baseline methods. The results are in Figure \ref{fig:sve_more_masafusion}. 





\end{document}